\documentclass[lettersize,10pt,twocolumn]{article}
\usepackage[utf8]{inputenc}
\usepackage{amsmath,graphicx}
\usepackage{amsfonts}
\usepackage{amssymb}
\usepackage{algorithm}
\usepackage{algpseudocode}
\usepackage[nice]{nicefrac}
\usepackage{url}
\usepackage{mathtools}
\usepackage{multirow}
\usepackage[table]{xcolor}
\usepackage{tabularx}
\usepackage[textwidth=1.145\textwidth,textheight=1.2\textheight]{geometry}
\usepackage{sectsty}
\usepackage{cleveref}
\usepackage{ulem}
\sectionfont{\normalsize}
\subsectionfont{\normalsize}

\title{\normalsize This is a preliminary version of FaceQSORT. The final version is currently under review. \\ \vspace{0.4cm}
\LARGE FaceQSORT: a Multi-Face Tracking Method based on Biometric and Appearance Features}
\author{\normalsize Robert Jöchl and Andreas Uhl\\\normalsize University of Salzburg, Department of Artificial Intelligence and Human Interfaces, \\\normalsize Salzburg, Austria\\
\ \normalsize \{robert.joechl, andreas.uhl\}@plus.ac.at}
\date{}

\begin{document}

\maketitle

\begin{abstract}
In this work, a novel multi-face tracking method named FaceQSORT is proposed. To mitigate multi-face tracking challenges (e.g., partially occluded or lateral faces), FaceQSORT combines biometric and visual appearance features (extracted from the same image (face) patch) for association. The Q in FaceQSORT refers to the scenario for which FaceQSORT is desinged, i.e. tracking people's faces as they move towards a gate in a Queue. This scenario is also reflected in the new dataset `Paris Lodron University Salzburg Faces in a Queue', which is made publicly available as part of this work. The dataset consists of a total of seven fully annotated and challenging sequences (12730 frames) and is utilized together with two other publicly available datasets for the experimental evaluation. It is shown that FaceQSORT outperforms state-of-the-art trackers in the considered scenario. To provide a deeper insight into FaceQSORT, comprehensive experiments are conducted evaluating the parameter selection, a different similarity metric and the utilized face recognition model (used to extract biometric features).
\end{abstract}

\section{Introduction}
\label{sec:intro}
Multi-face tracking is about finding the trajectories of all faces (identities) appearing in a sequence of images. This work focuses on tracking the people's faces while they move towards a gate in a queue.

The assumed scenario is that a group of people is moving towards a gate (e.g., to enter a sports stadium). As they move towards a single gate, they form a queue. However, as this is a leisure activity, this queue will not be very well organized (people chat with each other, they eat, they move around in a disorderly fashion, there may be pushing and shoving, etc.). In the assumed scenario, the main objective is to track people's faces as they move towards the gate. For this purpose, a camera is mounted on the gate. The camera is directed at the queue such that the first person visible is standing directly in front of the gate. However, since the queue is probably not very well organized, the main tracking challenges are occlusions (partially and full), out-of-plane rotations (lateral faces) and non-linear motion. In the assumed scenario, tracking people's faces can facilitate entrance by enabling people who have registered a face image during pre-enrollment to pass through the gate without additional checks. On the other hand, it can help to identify unwanted persons (i.e., persons banned from the premises) on their way to the gate and deny them entrance. Furthermore, each ticket is usually personalized. Tracking people's faces therefore enables a set of reference face images to be stored with the ticket, which assists in identifying individuals during the event.

In this work, a novel multi-face tracker named FaceQSORT is proposed. FaceQSORT is designed to track multiple face in the assumed scenario (i.e., people are moving towards a gate in a queue). For this purpose, two different types of features are extracted from the same modality (i.e., the detected face). The two features (i.e., biometric and appearance features) are combined into a single cost value, which forms the basis for association and is the core of FaceQSORT. To the best of our knowledge, FaceQSORT is the first multi-face tracking method that combines two features extracted from the same image patch. In general, multi-face trackers \cite{Zhao20a,Wang21a,Alayary22a,Tran21a,Marcetic18a,Barquero20a,Pernici20a} rely solely on biometric features. The combination of the two features helps to mitigate tracking challenges (e.g., lateral or partially occluded faces). FaceQSORT is evaluated based on a new multi-face tracking dataset. This dataset reflects the assumed scenario and is released as part of this work. To the best of our knowledge, the published dataset is the first multi-face tracking dataset in which people move in an (unorganized) queue towards a camera (gate). There is only one similar dataset, the ChokePoint dataset \cite{Yongkang11a} (people moving through a portal), which is also used for experimental evaluation. In the released dataset, however, a queue is simulated in which people are talking, eating, pushing, shoving, etc. The third dataset used for experimental evaluation is the MusicVideo dataset \cite{Zhang16b}. This dataset consists of unconstrained videos that do not reflect the considered scenario and is only utilized when comparing FaceQSORT with state-of-the-art trackers.

In summary, the main contributions of this work are:
\begin{itemize}
 \item A novel multi-face tracking method (FaceQSORT) is proposed that combines two different features (i.e., biometric features and appearance features) extracted from the same image (face) patch.
 \item A comprehensive experimental evaluation is performed on three datasets, comparing different face recognition models, analyzing parameter selection, applying a different similarity metric and comparing FaceQSORT with different state-of-the-art trackers.
 \item A dedicated multi-face tracking dataset is released. This dataset constitutes of seven annotated sequences (12730 frames) that reflect the assumed scenario (people moving in queue towards a gate).
 \item Annotations are provided for 6 sequences from the ChokePoint dataset.
\end{itemize}

The remainder of this paper is organized as follows: related work is listed in section \ref{sec:related_work} and the proposed FaceQSORT is described in section \ref{sec:faceqsort}. Section \ref{sec:dataset} provides an overview of the proposed dataset. The experimental setup is described in section \ref{sec:exp_setup}. In section \ref{sec:exp_eval} the experimental results obtained are reported and discussed. The key insights are summarized in the last section \ref{sec:con}.

\section{Related Work}
\label{sec:related_work}
To solve a general MOT problem, basically two paradigms exist: (i) joint-detection-association (JDA), and (ii) tracking-by-detection (TbD). With a JDA based tracker, basically everything is learned (end-to-end), i.e. the object detection, the relevant cues for the association and the association of objects across frames. Recent methods in this context are \cite{Yu23a,Zhou23a,Yang23a,Feng24a}. In this work, however, the focus is on TbD methods \cite{Dai19a,Fu19a,Bao21a,Zhang24a}. The first step of a tracker based on the TbD paradigm is the detection of all objects (e.g., faces) in the current frame. The tracking performance is therefore directly related to the detection performance. Tracking is then about finding bipartite matches (i.e., association problem) for all detected objects and active tracks (i.e., detected objects from previous frames). A common way to solve this association problem is to apply the Hungarian algorithm \cite{Kuhn55a} to a cost matrix (representing the costs of matching the $i$-th active track with the $j$-th detected object). These costs can, for example, be based on motion cues and/or appearance cues. Prominent examples for TbD trackers are Simple Online Realtime Tracking (SORT) \cite{Bewley16a} (motion based association), DeepSORT \cite{Wojke17a} (associated based on appearance features extracted by a Convolutional Neural Network (CNN)) and StrongSORT \cite{Du23a}. StrongSORT is an improved version of DeepSORT that comprises a more discriminative feature extractor \cite{Luo20a}, replaces the matching cascade with a vanilla matching, etc. A combination of motion and appearance models in a single network, called UMA, is proposed in \cite{Yin20a}. In \cite{Nodehi22a}, a fusion of six distance metrics (for motion and appearance cues) is used for association. One advantage of TbD trackers over JDA trackers is that pre-trained models can be utilized for the individual subtasks (e.g., a face detector and a face recognition model), while performance is similar (as demonstrated in \cite{Seidenschwarz23a}).

In TbD based multi-face trackers, association is usually based on the similarity of biometric features (i.e., features extracted by a face recognition model). For example, a two-stage association based on motion and biometric features (ArcFace \cite{Deng19a} trained on the MS-Celeb-1M dataset \cite{Guo16a}) is proposed in \cite{Zhao20a}. Multi-face tracking based on DeepSORT \cite{Wojke17a} is proposed in \cite{Wang21a}. The authors evaluate different combinations of face detectors (i.e., MTCNN \cite{Zhang16a}, SSD \cite{Liu16a} and R-FCN \cite{Dai16a}) and a face recognition model trained with two different loss functions (i.e., cosine softmax classifier loss \cite{Wojke18a} and angle softmax classifier loss \cite{Liu17a}). Biometric features of the upper facial regions (i.e., eyes, eyebrows and forehead) are used to track masked faces in \cite{Alayary22a}. A multi-face tracking method based on SORT \cite{Bewley16a} (i.e., motion based association) with a similarity matching block (comparing the stored with the currently computed biometric features, i.e. ArcFace \cite{Deng19a}) as a fallback for all unmatched detections showing a frontal face is proposed in \cite{Tran21a}. The method is named ReSORT, because IDs can be recovered based on the similarity matching block. Biometric features as fallback if motion based association fails is also proposed in \cite{Marcetic18a}. In \cite{Barquero20a}, a single object tracking method is used to predict the positions of an active track. The position-based matches are then corrected using biometric features. A multi-face tracking method based on biometric features only is presented in \cite{Pernici20a}. Cumulative learning of biometric features is performed based on a memory module, where selectively redundant features are removed. The proposed method is referred as IdOL (Identity Online Learning).

The biometric features used for tracking are usually based on pre-trained face recognition models (e.g., \cite{Deng19a,Parkhi15a,Schroff15a,Zhong21a}) that are typically trained on more or less frontal, unoccluded faces. In a multi-face tracking scenario, however, lateral faces (out-of-plane rotations) and partially occluded faces occur. To address this issue, a multi-modality tracker, Online Multi-face Tracking with Multi-modality Cascaded Matching (OMTMCM), is proposed in \cite{Weng23a}. The proposed method utilizes body and face features and includes two stages: (i) detection alignment, and (ii) detection association. The detected faces and bodies are aligned in the first stage (i.e., detection alignment). Detection association is performed in a cascade, whereby matching based on biometric features is performed only with those detections that could not be matched with body features. The biometric features are extracted by a SeNet \cite{Hu18a} pre-trained on the VGGFace2 database \cite{Cao18a} and the body features by a ResNet \cite{He16a} pre-trained on the MovieNet database \cite{Huang20a}. However, a corresponding body is not always visible to every detected face, especially in the assumed scenario in which people are moving towards a gate in a queue.

\begin{figure*}[ht!]
\centering
\includegraphics[width=0.84\linewidth]{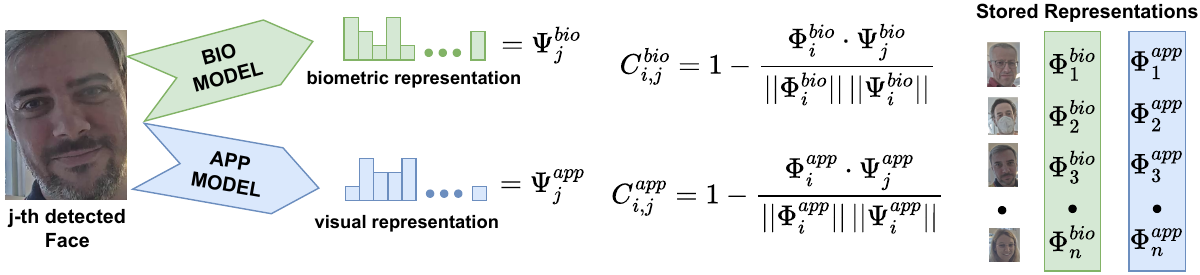}
\caption{The core of FaceQSORT. Biometric and appearance features are extracted from the same image (face) patch and the cosine similarity is calculated with all stored representations (features) from the active track memory $\Phi$.}
\label{fig:FaceQSORT}
\end{figure*}

\section{FaceQSORT}
\label{sec:faceqsort}
FaceQSORT is a TbD-based tracker designed to track the Faces of multiple people as they move towards a gate in a Queue (source code is made available\footnote{https://gitlab.cosy.sbg.ac.at/wavelab/faceqsort}). SORT stands for Simple Online Realtime Tracking and illustrates that FaceQSORT is inspired by StrongSORT \cite{Du23a}. The core of FaceQSORT is the combination of two different types of features (i.e., biometric and appearance features) extracted from the same image (face) patch (i.e., the Bounding Box (BBox) returned by the face detector). By using two different features from the same modality, the required presence of body and face (as utilized in \cite{Weng23a}), which hardly exists in the considered scenario, is eliminated. Furthermore, if only a single modality is utilized, the need for a second detector and some alignment overhead is avoided.

Biometric features, extracted by a face recognition model, provide a biometric representation of the face. These features should be distinct for each identity. However, due to tracking challenges (e.g., partially occluded, lateral or deformed faces), the distinctiveness of these features might be limited. For this reason, biometric features are combined with appearance features. Appearance features, extracted by a generic image classifier, provide a general visual representation of the image (face) patch and the combination with biometric features should mitigate multi-face tracking challenges. The combination of these features is performed on the cost matrix level and is depicted in Figure \ref{fig:FaceQSORT}.

In common TbD based methods, the association relies on a cost matrix $C^{|\Phi| \times |\Psi|}$, where $\Phi$ is the set of active tracks (active track memory) and $\Psi$ the set of detected objects. An element $C_{i,j}$ represents the cost of matching the $i$-th active track to the $j$-th detected object. The main objective of association is to find bipartite matches where the global cost is minimum (i.e., usually solved by the Hungarian algorithm \cite{Kuhn55a}). In principle, $C_{i,j}$ is based on a distance $d$ from a stored/predicted cue of the $i$-th active track to a calculated/observed cue of the $j$-th detected object. In FaceQSORT, the combination of biometric and appearance features is performed as follows,
\begin{equation}
 \begin{aligned}
   & \multicolumn{2}{@{}l}{\text{$C_{i,j}^{app/bio} = \lambda C_{i,j}^{bio} + (1-\lambda) C_{i,j}^{app},$}} \\
    \text{with} & \quad \text{$C_{i,j}^{bio} = d_{bio}(\Phi_{i}^{bio},\Psi_{j}^{bio}),$} \\
    \text{and}  & \quad \text{$C_{i,j}^{app} = d_{app}(\Phi_{i}^{app},\Psi_{j}^{app}).$}
  \end{aligned}
\label{eq:Cij_app_bio}
\end{equation}
The distance measure $ d_{bio}$ and $ d_{app}$ is the cosine similarity (see Figure \ref{fig:FaceQSORT}) of the stored and observed features and  $\lambda \in \mathbb{R}_{[0,1]}$ weights (scales) the resulting costs (distances). The combination of biometric and appearance is unique (to the best of our knowledge) and represents the core of FaceQSORT.

Similar to StrongSORT a single cost matrix element is gated by a maximum spatial distance $\theta_{pos}$, i.e.
\begin{equation}
C_{i,j}^{app/bio}  = \begin{cases}
\text{inf} & d_{pos}(\Phi_{i}^{pos},\Psi_{j}^{pos}) > \theta_{pos} \\
 C_{i,j}^{app/bio} & \text{otherwise}
 \label{eq:Cij_app_bio_threshold}
\end{cases},
\end{equation}
where $d_{pos}$ is the Mahalanobis distance. This is done to avoid uncertain matches. The spatial distance costs, $C_{i,j}^{pos} = d_{pos}(\Phi_{i}^{pos},\Psi_{i}^{pos})$, are also integrated into the final cost value, i.e.
\begin{equation}
  C_{i,j} = \beta C_{i,j}^{app/bio}+(1-\beta) C_{i,j}^{pos},
  \label{eq:Cij}
\end{equation}
and a general threshold $\theta$ is applied, i.e.
\begin{equation}
C_{i,j} = \begin{cases}
\text{inf} & C_{i,j} > \theta \\
 C_{i,j} & \text{otherwise}
 \label{eq:Cij_threshold}
\end{cases}.
\end{equation}
Based on the cost matrix $C$, the bipartite matching problem is formulated as follows,
\begin{equation}
 \begin{aligned}
    \multicolumn{2}{@{}c}{\text{$argmax_{\alpha_{i,j} \in \{0,1\}} \sum_{i=1}^{|\Phi (t)|} \sum_{j=1}^{|\Psi (t)|} \alpha_{i,j} (C_{max} - C_{i,j})$}} \\
    \text{s.t.} & \begin{cases}
\sum_{i} \alpha_{i,j} \leq 1, & \forall_{j} = 1, \dots, |\Psi(t)| \\
\sum_{j} \alpha_{i,j} \leq 1, & \forall_{i} = 1, \dots, |\Phi(t)| \\
\end{cases}
  \end{aligned},
  \label{eq:matching}
\end{equation}
where $C_{max}$ denotes the largest element in $C$ (less than inf). The problem is solved by the Hungarian algorithm \cite{Kuhn55a}.

Unlike StrongSORT, the matching cascade proposed with DeepSORT \cite{Wojke17a} is reactivated. In the matching cascade, matching is performed in sequence, starting with the tracks that were correctly matched in the previous frame and ending with the tracks that have not been matched for the longest time. The matching cascade reduces the search space and could be beneficial when long occlusions are allowed. For all unmatched detections (after the matching cascade), an IoU-based association (as proposed in SORT \cite{Bewley16a}) is performed as a fallback routine.

The active track memory $\Phi$, stores a unique ID, the biometric features $\Phi^{bio}$, the appearance features $\Phi^{app}$ and the position $\Phi^{pos}$ for each active track. A track becomes inactive (is deleted from $\Phi$) if it could not be matched with a detected face in $N_{max}$ consecutive frames. When a new track is added to $\Phi$, it is in a tentative state and is only confirmed if the track could be matched in the next $N_{init}$ frames. Otherwise, the tentative track is deleted from $\Phi$. Similar to StrongSORT, an Exponential Moving Average (EMA) \cite{Wang20a} approach is used to update the stored biometric and appearance features from the $i$-th matched track, i.e.
\begin{equation}
  \Phi_i^t = \alpha \Phi_i^{t-1} + (1-\alpha)\Psi_i^t,
  \label{eqn:memory_upate}
\end{equation}
where $\Psi_i^t$ represents the extracted features from the current frame $t$ of the face associated with track $i$ and the parameter $\alpha$ is a momentum term. As stated in \cite{Du23a}, this updating strategy leverages the information of inter-frame feature changes and can mitigate detection noises. $\Phi$ is maintained separately for biometric and appearance features (i.e., $\Phi^{bio}$ and $\Phi^{app}$). The prediction of the position of the $i$-th active track ($\Phi_{i}^{pos}$) in the next frame is based on the NSA Kalman filter \cite{Du21a}. The general steps that are performed in FaceQSORT for each frame are summarized in Algorithm \ref{alg:FaceQSORT}.

\begin{algorithm}
\caption{FaceQSORT at frame $t$}\label{alg:FaceQSORT}
\begin{algorithmic}[1]
\Statex \textbf{Input:} $\Phi$ and $\Psi$
\State predict position $\Phi^{pos}$ (Kalman Filter)
\State $\Psi^{\prime} = \Psi$; $\Phi^{+} \subseteq \Phi$ (confirmed tracks)
\For{$i = 1, \dots, max(\Phi^{+age})$}
\If{$|\Psi^{\prime}| > 0$}
\State calculate $C$ by Eqn. (\ref{eq:Cij}) using $\Phi^{+ age=i}$ and $\Psi^{\prime}$
\State find matches Eqn. (\ref{eq:matching})
\State update $\Psi^{\prime}$ (unmatched detections)
\Else
\State break
\EndIf
\EndFor
\State IoU fallback matching if $|\Psi^{\prime}| > 0$
\State update\//add $\Phi^{bio}$ and $ \Phi^{app}$ for all detections (Eqn. (\ref{eqn:memory_upate}))

\end{algorithmic}
\end{algorithm}

\subsection{Computational Complexity}
As shown in Algorithm \ref{alg:FaceQSORT}, FaceQSORT essentially comprises the matching cascade, the IoU fallback matching, the feature updating and the position prediction. In general, the complexity of finding bipartite matches using the Hungarian algorithm is $O(|\Phi|^{2} \times |\Psi|)$. This is also the worst case with cascade matching. In the considered scenario, it is very likely that the detected face $\Psi$ in frame $t$ were also present in frame $t-1$. Thus, it is reasonable that the complexity can be reduced by the matching cascade (i.e., to $O((\nicefrac{|\Phi|}{x})^{2} \times \Psi)$). For all unmatched detections after the matching cascade, the complexity of the IoU fallback matching is in the worst case again $O(|\Phi|^{2} \times |\Psi|)$. The complexity for updating the memory for biometric and appearance features ($\Phi_{bio}$ and $ \Phi_{app}$) is $O(|\Psi|)$ in each case. Finally, predicting the positions in frame $t+1$ using the Kalman filter has basically a complexity of $O(|\Phi|)$. Thus, the overall complexity of FaceQSORT is,
\begin{equation}
 O(|\Phi|^{2} \times |\Psi| + |\Phi|^{2} \times |\Psi| + 2|\Psi| + |\Phi|),
\end{equation}
which corresponds to the complexity class $O(|\Phi|^{2} \times |\Psi|)$.

\section{PLUS Faces in a Queue Dataset}
\label{sec:dataset}
The Paris Lodron University Salzburg Faces in a Queue (PLUSFiaQ) dataset is a new multi-face tracking dataset consisting of a total of seven different sequences. This dataset is made publicly available\footnote{\url{https://www.wavelab.at/sources/plusfiaq}}. An overview of the individual sequences is given in Table \ref{tab:seq_details}. In total, the seven sequences comprise 12730 fully annotated frames. This corresponds to about 8 minutes and 30 seconds of video material (25 frames per second). The recorded sequences reflect the assumed scenario, in which several people move towards a gate in a queue (e.g., to enter a sports stadium). Practical use cases for tracking peoples faces as they walk towards a gate are: (i) entrance can be facilitated by registering a face image with the ticket and enabling the person (if correctly recognized) to pass through the gate without additional checks, (ii) potentially unwanted persons (i.e., persons banned from the premises) can be identified on the way to the gate and prevented from entering, (iii) for each ticket presented (which is usually personalized), a set of reference face images can be stored to help identify individuals during the event. Since the time and the area in which a person is present in the visible range of the camera (mounted on the gate) is very limited, the tracking should be highly accurate. In addition, inter-person tracking ID (trID) switches are crucial in the considered scenario. Informed consent was obtained from all participants for all released sequences.

\begin{table}
\begin{center}
\caption{Sequence (SEQ) details, where F denotes the frames, P the individual person and $\text{O}_d$ the occlusion duration. }
\label{tab:seq_details}
\scriptsize
\begin{tabular}{ c | c | c | c | c | c | c  }
SEQ-ID &  \# F & \# P & $\mu(\text{O}_d)$ & $\sigma(\text{O}_d)$ & min($\text{O}_d$) & max($\text{O}_d)$\\ \hline \hline
01 & 1774 &   3  & 52.00 & 32.53 & 29 & 75   \\
02 & 3051 &  12  & 44.52 & 55.30 & 1 & 500   \\
03 &  701 &  12  & 42.04 & 50.38 & 1 & 250   \\
04 &  576 &  10  & 38.26 & 38.57 & 2 & 206   \\
05 & 3126 &  11  & 39.35 & 41.88 & 3 & 271   \\
06 & 1301 &  12  & 43.68 & 39.17 & 4 & 197   \\
07 & 2201 &  10  & 43.50 & 45.97 & 1 & 257   \\
\hline \hline
\end{tabular}
\end{center}
\end{table}

\subsection{Scenario Details}
\begin{figure}[!t]
\centering
\includegraphics[width=0.34\linewidth]{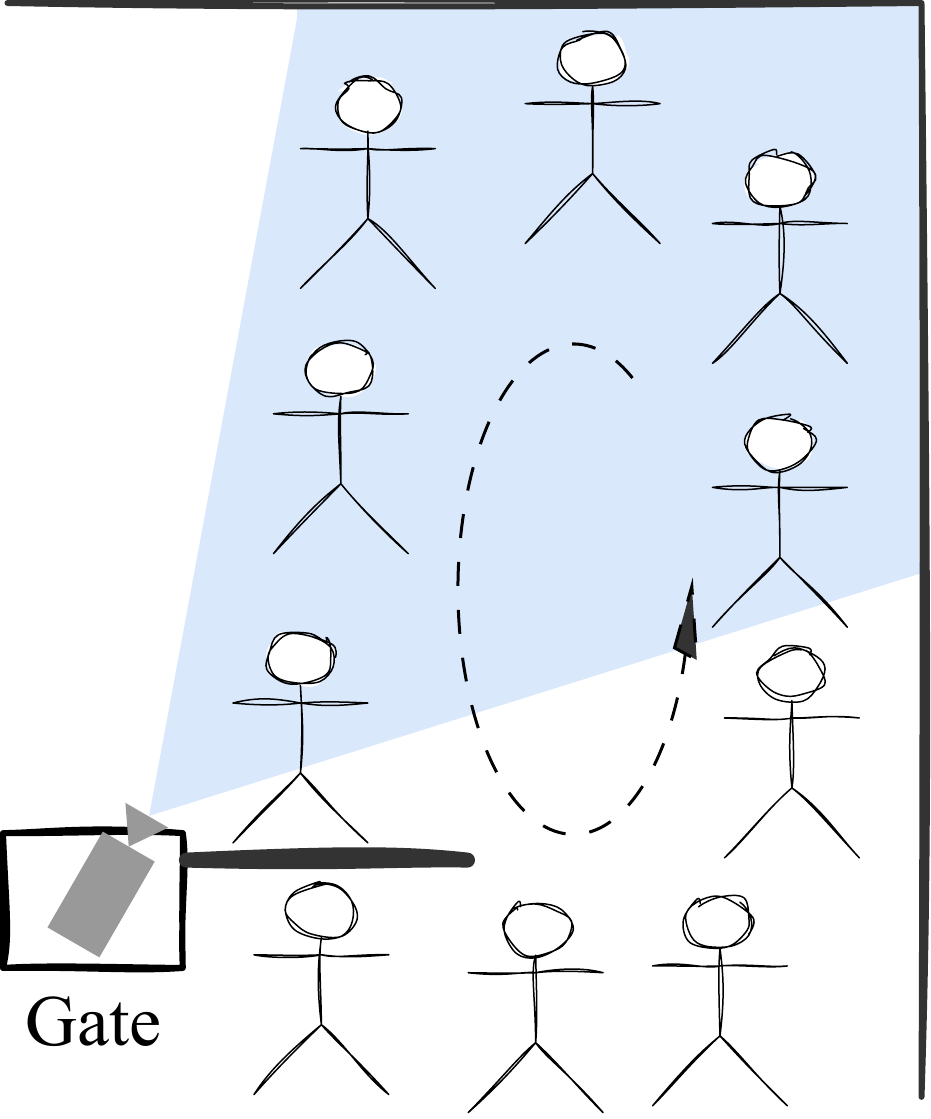}
\caption{A schematic depiction of the recorded scenario. The area shaded in blue represents the visible area of the camera.}
\label{fig:schemantic_scenario}
\end{figure}

\begin{figure*}[!t]
\centering
\begin{minipage}[b]{0.134\textwidth}
\includegraphics[width=0.98\linewidth]{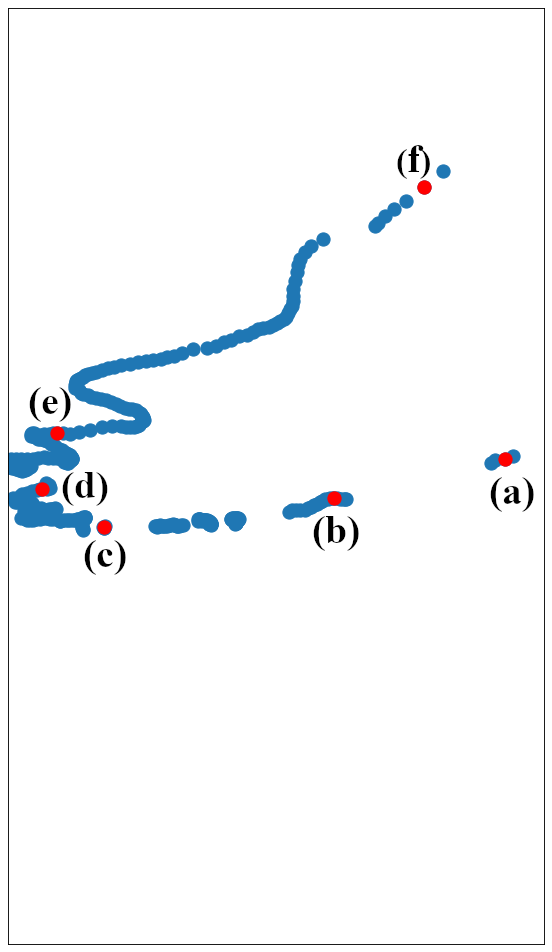}
\centerline{trajectory}\medskip
\end{minipage}
\begin{minipage}[b]{0.13\textwidth}
\includegraphics[width=0.98\linewidth]{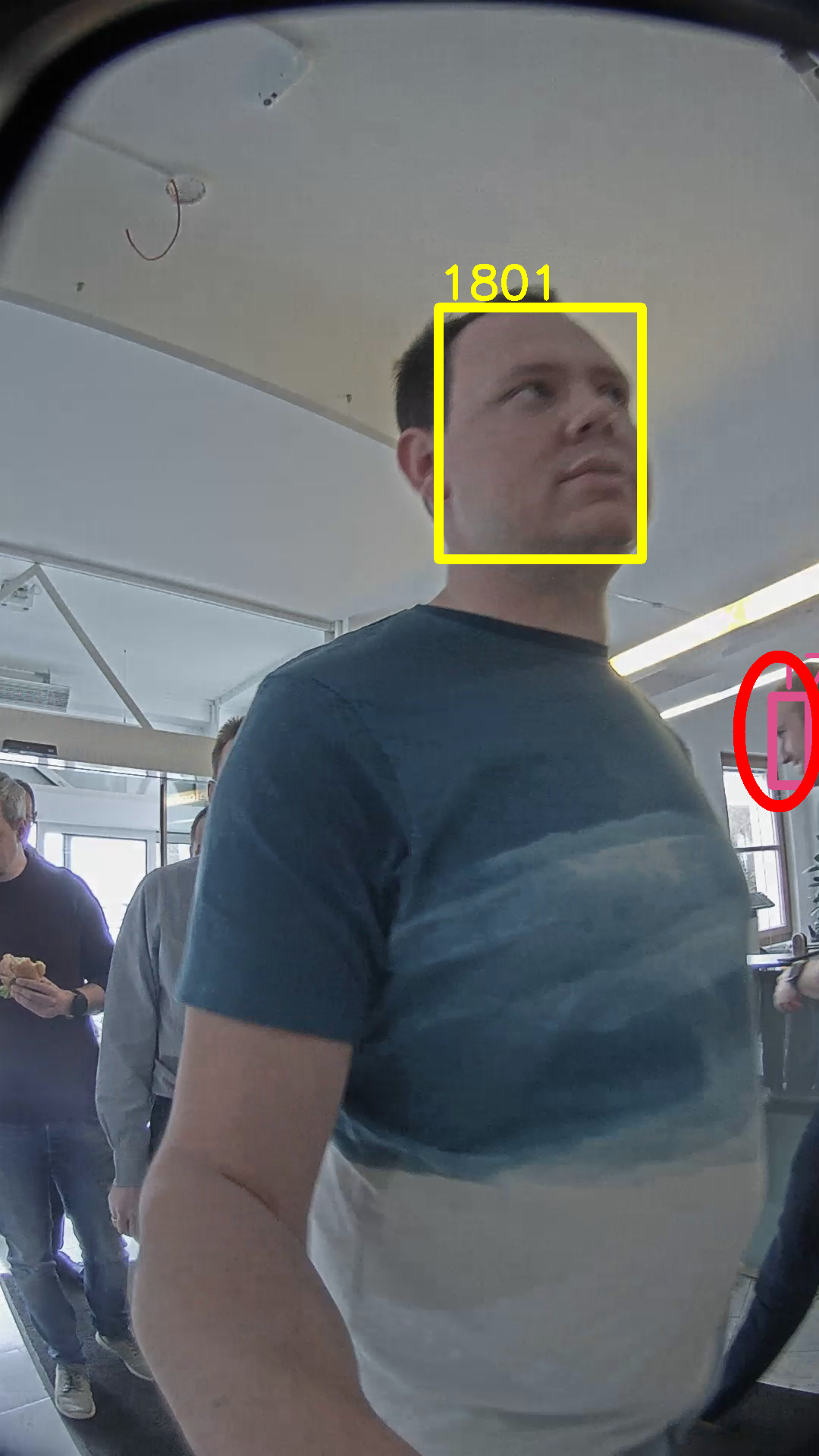}
\centerline{(a)}\medskip
\end{minipage}
\begin{minipage}[b]{0.13\textwidth}
\includegraphics[width=0.98\linewidth]{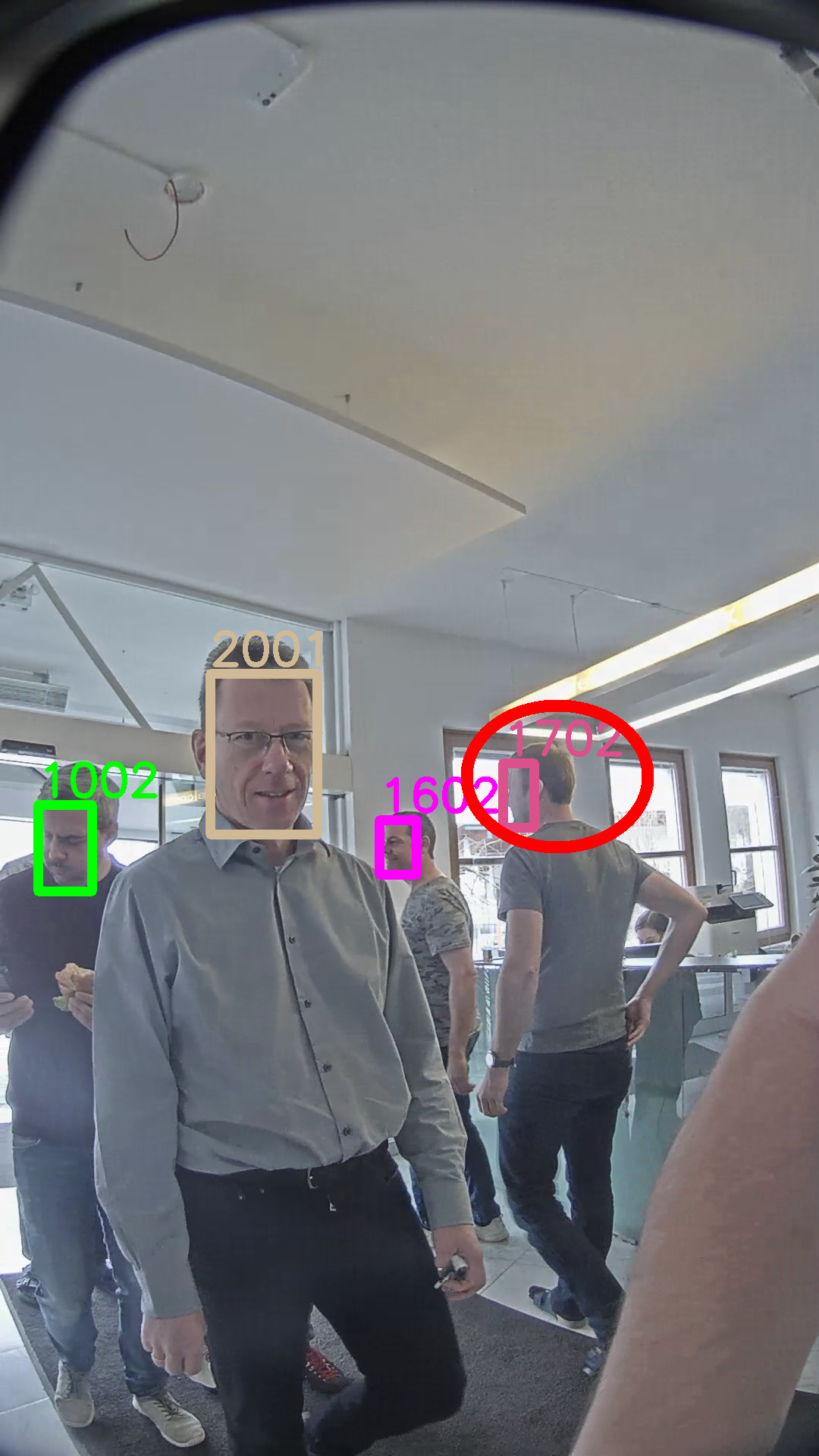}
\centerline{(b)}\medskip
\end{minipage}
\begin{minipage}[b]{0.13\textwidth}
\includegraphics[width=0.98\linewidth]{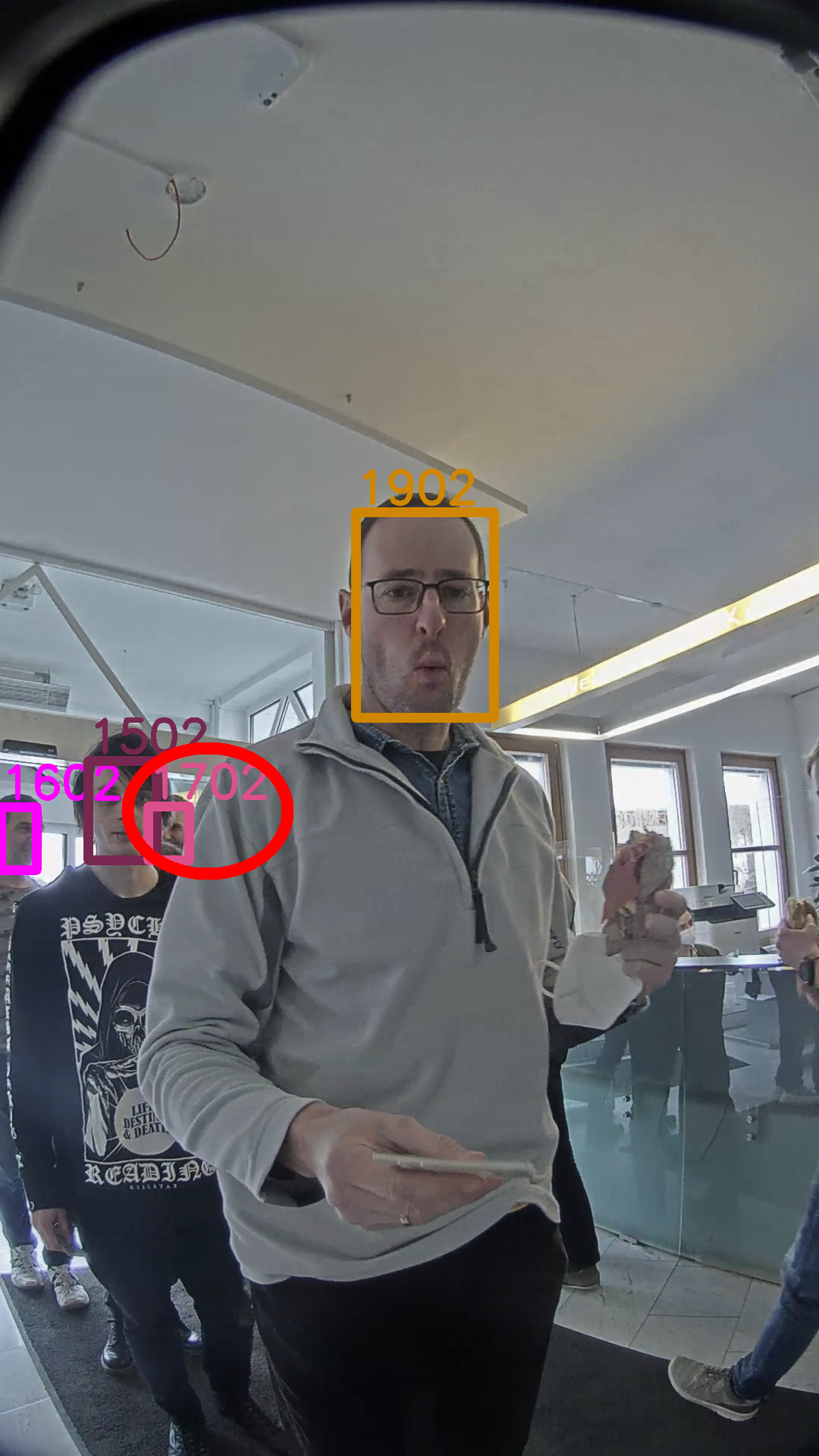}
\centerline{(c)}\medskip
\end{minipage}
\begin{minipage}[b]{0.13\textwidth}
\includegraphics[width=0.98\linewidth]{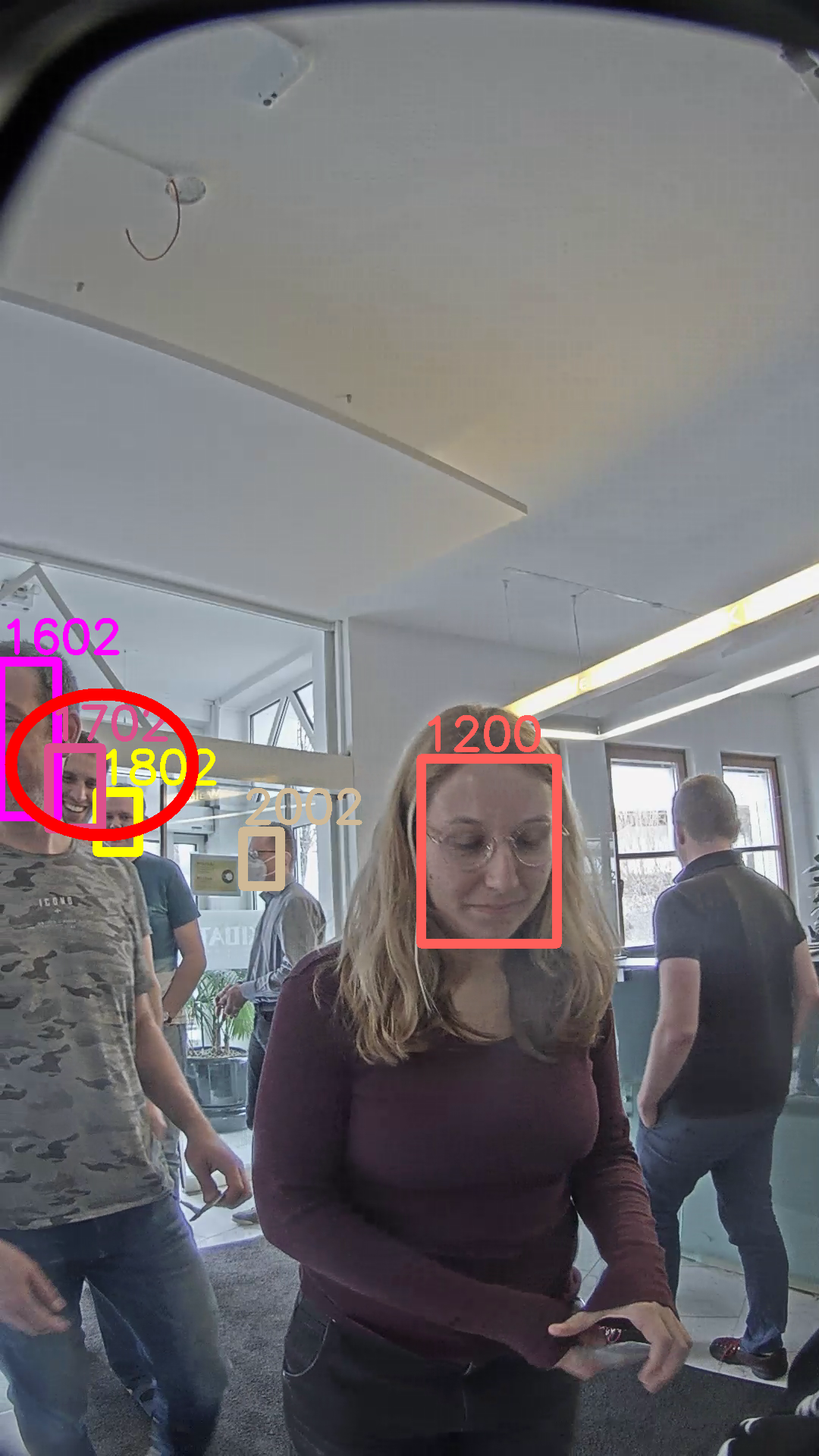}
\centerline{(d)}\medskip
\end{minipage}
\begin{minipage}[b]{0.13\textwidth}
\includegraphics[width=0.98\linewidth]{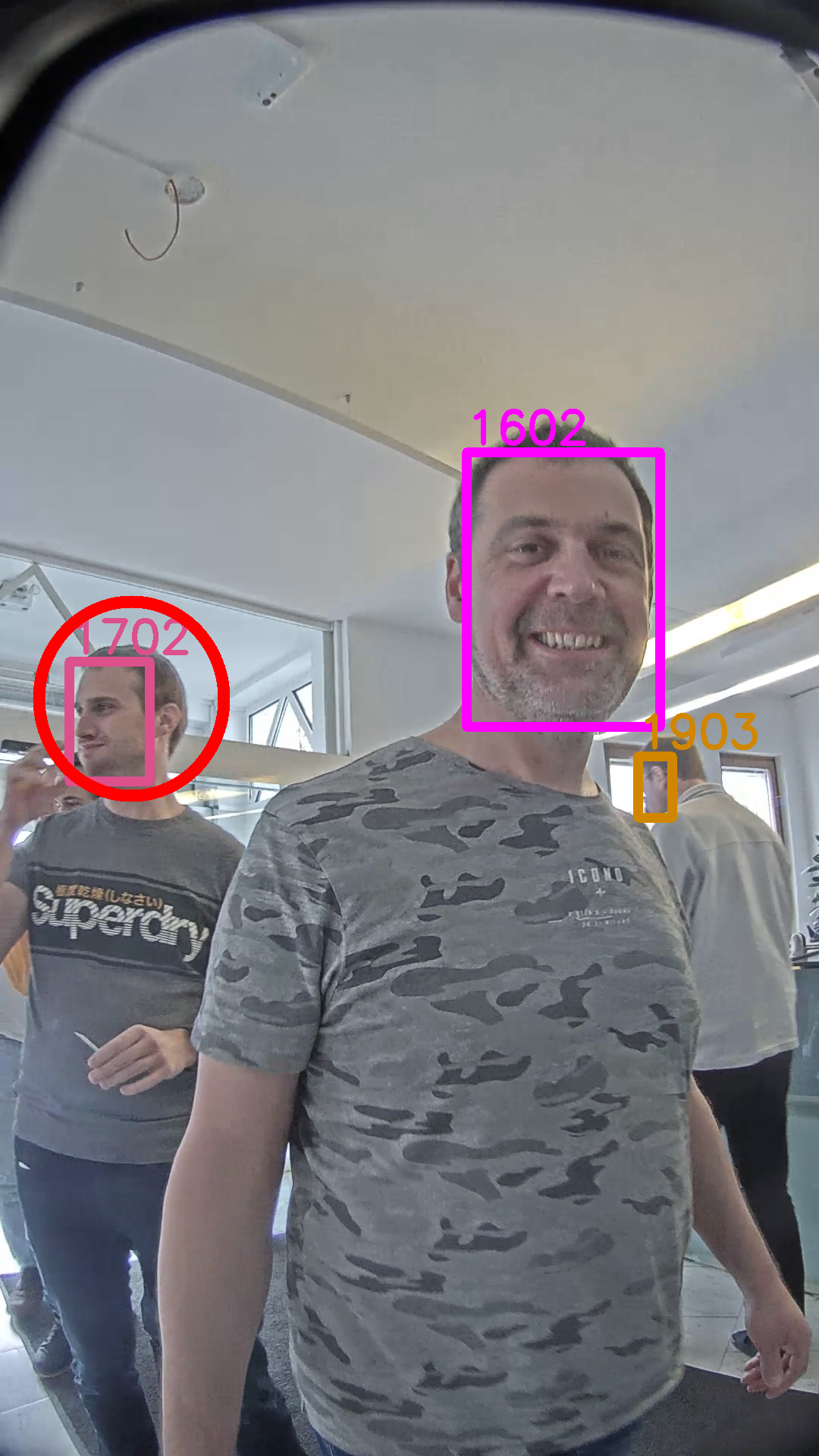}
\centerline{(e)}\medskip
\end{minipage}
\begin{minipage}[b]{0.13\textwidth}
\includegraphics[width=0.98\linewidth]{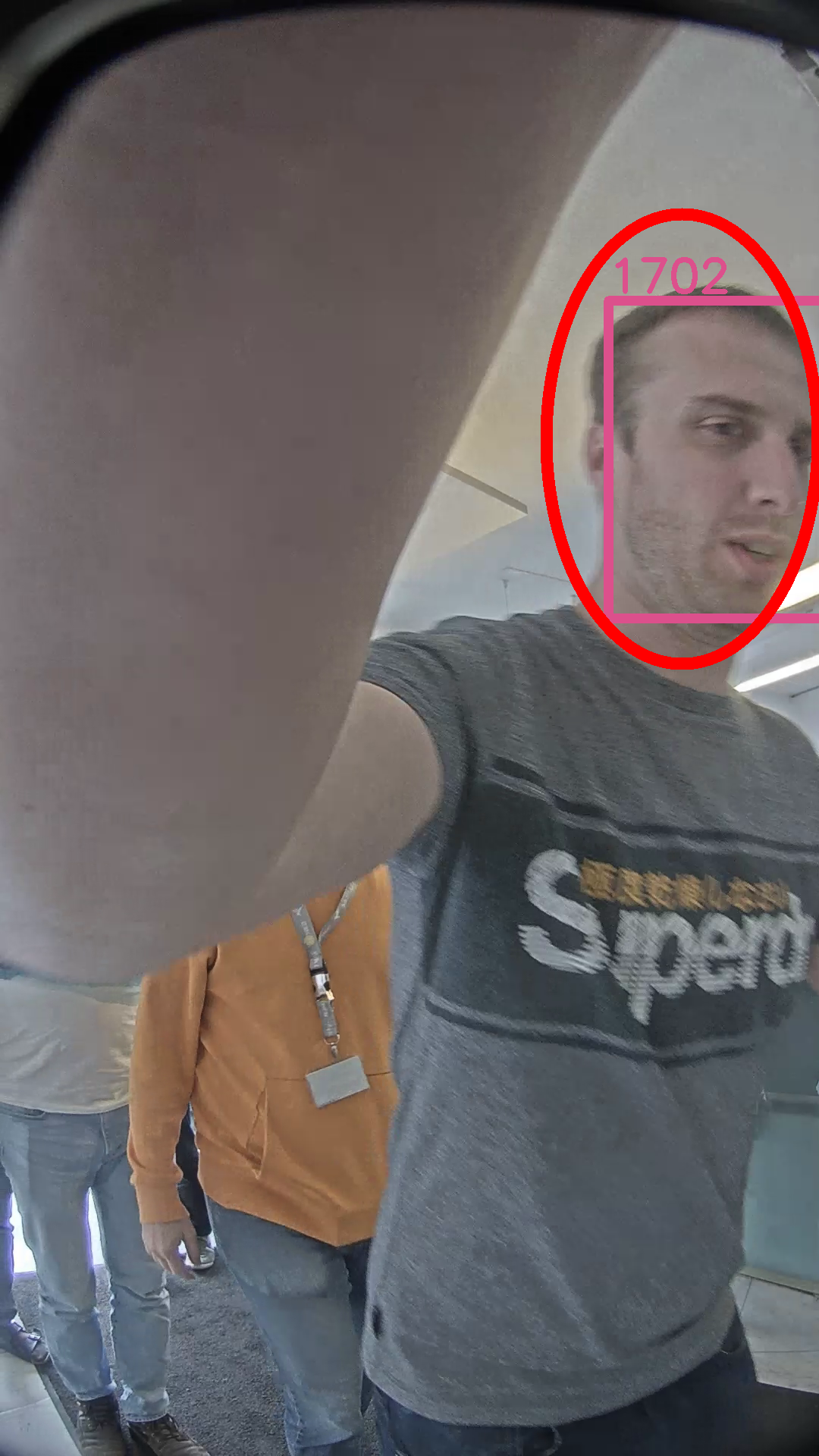}
\centerline{(f)}\medskip
\end{minipage}
\caption{Example of a trajectory of the person with trID 1702.}
\label{fig:trajectory_example}
\end{figure*}

The assumed scenario (i.e., queue in front of a gate) is simulated with 12 different people. These persons move towards a camera equipped gate. On their way to the gate, they form a queue. However, this queue is not very well organized, i.e. people are chatting, eating, moving quickly, walking backwards, pushing or shoving. As soon as a person has entered the gate, they leave the visible area of the camera and thus the scene. To simulate multiple runs, each person goes more than once through the gate. For this reason, the people move in a counterclockwise circle (as depicted in Figure \ref{fig:schemantic_scenario}). Thus, when a person re-enters the scene, the person first moves away from the gate (camera). This is also illustrated in Figure \ref{fig:trajectory_example}, where the trajectory of the top-left Bounding-Box coordinates from a specific person is shown (i.e., trID 1702, marked by a red circle). At position (a) the person re-enters the scene, then moves away from the gate (b) until he is roughly farthest away at (c) and then moves towards the gate again at (d) and (e). Finally, the person again leaves the scene (goes through the gate) at position (f). This loop is repeated several times.

The only exception is sequence 1, in which only three people are involved, and they move clockwise, i.e. when they re-enter the scene, they move directly towards the gate. This sequence is also the most simple sequence.

\subsection{Annotation Details}
Annotations provided comprise a Bounding-Box (BB) for each face, a corresponding tracking identity (trID) and a target class flag. In addition, a visibility measure and a ‘next at gate’ flag are included, but these have not been fully adjusted and corrected at this stage. Furthermore, the alignment of the BBs is not defined, thus the contained areas of the face can differ between the BBs.

The trID is a four-digit number (e.g., xxyy), where the first two digits (xx) represent the personal identifiers and the last two digits (yy) represent a consecutive tracking number (per sequence). A new tracking number is assigned when a person has left the scene (goes through the gate) and re-enters the scene again. For example, the person in figure \ref{fig:trajectory_example} (a), who is directly in front of the gate, has the personal identifier 18. The current tracking number is 1 (i.e., trID 1801). However, after leaving and re-entering the scene, the tracking number is incremented by one (i.e., trID 1802), as shown in Figure \ref{fig:trajectory_example} (d). For all non-target class objects (faces) a personal identifier of 99 is assigned. The individual non-target class instances are assigned a dedicated tracking number, independent of how often they have re-entered the scene.

For annotation, the VGG Image Annotator (VIA)\footnote{http://www.robots.ox.ac.uk/ vgg/software/via} \cite{Dutta19a} was used. The corresponding exported annotation file (json file) is provided for each sequence. Furthermore, a Ground-Truth (GT) file according to the MOT20 format \cite{MOT20} is provided additionally, i.e. a csv file where each line represents one object instance and contains 9 values:
\begin{center}
   \textit{\small $<$frame ID$>$,$<$trID$>$,$<$BB left$>$,$<$BB top$>$,$<$BB width$>$,$<$BB height$>$,$<$conf$>$,$<$class$>$,$<$visibility$>$}
\end{center}
In case of ground-truth (GT), the 7$^{th}$ value (detector confidence score) acts as flag whether the entry is to be considered (i.e., 1: target class, 0: ignore). Since only multiple object (face) tracking is considered, the class label is constantly set to 1. Visibility represents the ratio how much of that object is visible.

As a starting point for the annotation, the BBs detected by yolov5\footnote{https://github.com/ultralytics/yolov5} and the trIDs predicted by StrongSORT \cite{Du23a} were used. These BBs and trIDs were then refined manually.

\subsection{Tracking Challenges}
\begin{figure}[!t]
\centering
\begin{minipage}[b]{0.32\linewidth}
\includegraphics[width=0.88\linewidth]{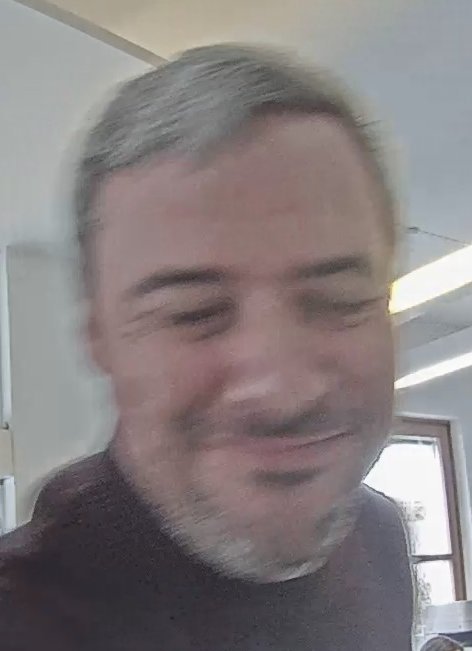}
\centerline{motion blur}\medskip
\end{minipage}
\begin{minipage}[b]{0.32\linewidth}
\includegraphics[width=0.855\linewidth]{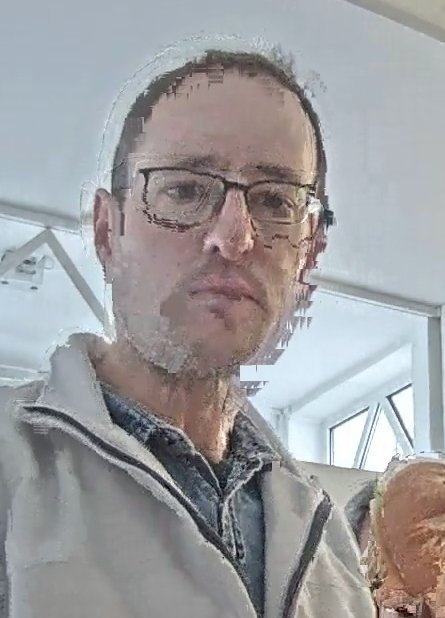}
\centerline{comp. artefacts}\medskip
\end{minipage}
\begin{minipage}[b]{0.32\linewidth}
\includegraphics[width=0.858\linewidth]{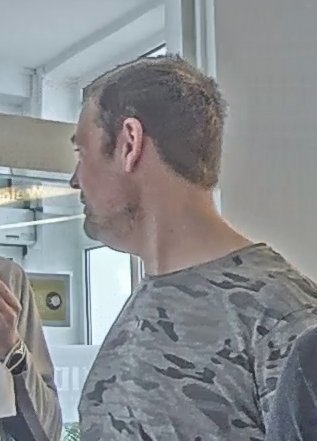}
\centerline{out-of-plane rot.}\medskip
\end{minipage}
\caption{Examples of tracking challenges.}
\label{fig:tracking_challenge_examples}
\end{figure}

The PLUSFiaQ dataset comprises various different multi-face tracking challenges. These challenges include: (i) fast motion, (ii) looking down or sidewards, (iii) cover the face (e.g., wear a mask), (iv) motion blur (see Figure \ref{fig:tracking_challenge_examples}), (v) deformations (i.e., eating or grimacing), (vi) compression artefacts (see Figure \ref{fig:tracking_challenge_examples}) and (vi) out-of-plane rotations (see Figure \ref{fig:tracking_challenge_examples}). However, the most difficult challenge are likely occlusions.

\begin{figure}[!t]
\centering
\includegraphics[width=0.88\linewidth]{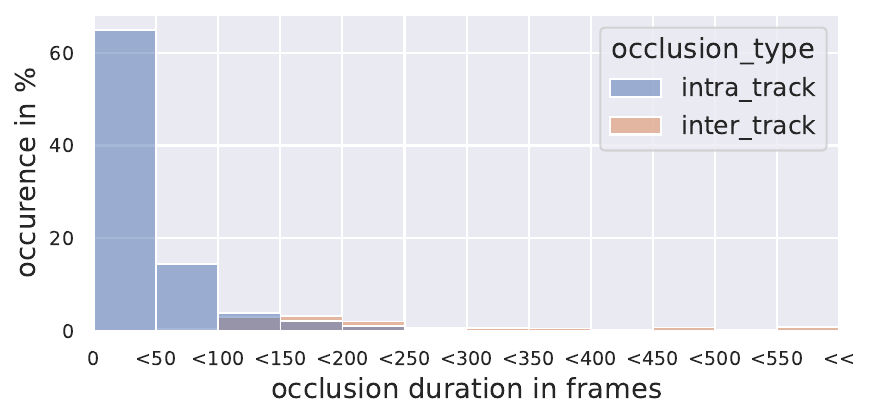}
\caption{Distribution of occlusion duration. Inter-track occlusions represent the time it takes (in frames) for a person to leave and re-enter the scene.}
\label{fig:occlusion_dist}
\end{figure}

In Figure \ref{fig:occlusion_dist}, the distribution of occlusion duration in frames is illustrated. As can be seen, around 80\% of all occlusions are between 1 and 100 frames in length. Inter-track occlusions are those occlusions that occur when a person has left the scene (goes through the gate, the tracking number is incremented by one, e.g., 1801 to 1802), and the inter-track occlusion duration is the time from leaving to re-entering the scene. On average, inter-track occlusions are 257.131 frames long, with maximum occlusion of 1400 frames. The average occlusion duration (intra-track) per sequence as well as minimum and maximum occlusion duration and the standard deviation can be seen in Table \ref{tab:seq_details}. In principle, when referring to occlusions, intra-track occlusions are meant. For example, the person with trID 1702 (shown in Figure \ref{fig:trajectory_example}) is occluded 11 times on the way from entering to leaving the scene, with a maximum occlusion duration of 83 frames and a minimum of 3 frames. The average number of occlusions per trID is 0.18, 5.91, 2.42, 3.07, 4.84, 4.89 and 4.03 for sequences 1 to 7, respectively.

\section{Experimental Setup}
\label{sec:exp_setup}
FaceQSORT is evaluated based on the proposed PLUSFiaQ (see section \ref{sec:dataset}), the ChokePoint \cite{Yongkang11a} and the MusicVideo \cite{Zhang16b} dataset. However, the MusicVideo dataset is only utilized when comparing FaceQSORT with state-of-the-art trackers. In the ChokePoint dataset, people moving through a portal (chokepoint) are recorded. The dataset comprises sequences from 18 different runs (i.e., different portal, session or walking direction), with 3 different camera perspectives available for each run (i.e., $18 \times 3 = 54$ available sequences). Of these 54 sequences, only 6 sequences show multiple persons per frame, which is similar to the assumed scenario in this paper. These 6 sequences are therefore used for evaluation. However, no annotations are available for these 6 sequences and are created accordingly. In the ChokePoint dataset, people are reflected in the open glass door as they walk through the portal. These reflections are annotated as non-target classes. The annotations created have the same format as described in section \ref{sec:dataset} and are made available along with the PLUSFiaQ dataset.

\subsection{Tracker Implementation}
FaceQSORT is carefully implemented according to the description in section \ref{sec:faceqsort}. The parameter $\alpha$ for the EMA feature updating strategy is set to $\alpha = 0.9$. The weighting parameter $\beta$ (for weighting the contribution of biometric and appearance similarity compared to spatial distance) is set to 0.98. This gives the biometric and appearance similarity the most weight. $\theta_{pos}$ is set to 9.4877 (the 0.95 quantile of the chi-squared distribution with 4 degrees of freedom). The general cost threshold $\theta$ is set to 0.2 and the minimum IoU (fallback routine) to 0.3. These parameters are set according to \cite{Du23a}. For all experiments, $N_{init}=1$ and the maximum age parameter is set to $N_{max}=100$ frames (i.e., all tracks that were not successfully matched for more than 100 frames are deleted). With this setting, most occlusions are taken into account while avoiding interference from inter-track occlusions (as shown in Figure \ref{fig:occlusion_dist}). As face detector yolov8\footnote{https://github.com/ultralytics/ultralytics} with a confidence score threshold of 0.4 is used.

\subsection{Evaluated Features}
To extract the appearance features, a generic object classification model is used, i.e. a ResNet \cite{He16a} trained on the ImageNet dataset \cite{Deng09} (resenet18 imported from torchvision models). The appearance features are combined with several different face (biometric) features. The python framework DeepFace \cite{Serengil20a,Serengil21a} provides a wrapper for multiple state-of-the-art face recognition models, i.e. VGG-Face (BIO-VG01) based on the VGG-Very-Deep-16 CNN architecture \cite{Parkhi15a} and evaluated on the Faces in the Wild \cite{Huang08a} and the YouTube Faces \cite{Wolf11a} dataset, Facenet \cite{Schroff15a} (BIO-FN01) trained on the CASIA-WebFace \cite{Yi14a} and the VGGFace2 \cite{Cao18a} database, OpenFace \cite{Amos16a} (BIO-OF01) trained on the CASIA-WebFace \cite{Yi14a} and FaceScrub \cite{Ng14a} dataset (based on paper), DeepFace \cite{Taigman14a} (BIO-DF01) trained on a large collection of Facebook images (i.e., the Social Face Classification (SFC) dataset), DeepID \cite{Sun14a} (BIO-DI01) trained on the Faces in the Wild \cite{Huang08a} dataset and tested, ArcFace \cite{Deng19a,Leondgarse22a} (BIO-AF01), SFace \cite{Zhong21a} (BIO-SF01) trained on CASIA-WebFace \cite{Yi14a} VGGFace2 \cite{Cao18a} and MS1MV2 and Dlib \cite{Davisking22a} (BIO-DL01) trained on the VGG-Face \cite{Parkhi15a} and FaceScrub \cite{Ng14a} dataset (plus a large number of images the author scraped from the internet). All these different face recognition models provided by DeepFace are evaluated. DeepFace also provides a 512 dimensional face descriptor for Facenet (BIO-FN02) and ArcFace (BIO-AF02) is also evaluated using different weights\footnote{https://www.digidow.eu/f/datasets/arcface-tensorflowlite/model.tflite}. In addition, the face recognition model used in \cite{Weng23a}, SeNet \cite{Hu18a} (BIO-SN01), is evaluated.

\subsection{Evaluation Metrics}
The Multiple Object Tracking Accuracy (MOTA)\cite{Bernardin08a} is a commonly used metric for evaluating multiple object tracking methods. With MOTA, tracking errors are accumulated on a frame-by-frame basis, i.e. wrongly detected objects (False Positive (FP)), missed objects (False Negative (FN)) and identity switches (IDSWs) are counted for each frame ($t$), summed up and normalized by the total amount of objects in the ground-truth (GT),
\begin{equation}
 MOTA = 1-\frac{\sum_{t}(FP_{t}+FN_{t}+IDSW_{t})}{|GT|}.
\end{equation}
An IDSW occurs if the ID assigned in the previous frame ($t-1$) differs from the ID assigned in the current frame ($t$). As MOTA only considers the previous frame, the entire tracking performance of an object during its lifetime is not taken into account. For example, if an object is tracked correctly for 90\% of its lifetime but yields the same number of IDSW as an object that is only tracked correctly for 60\%, the same MOTA score is achieved. Furthermore, the MOTA score can be dominated by the detector performance (i.e., TP and FP). Since the face detection is fixed for all experiments, only the normalized IDSW is calculated, i.e.

\begin{equation}
 IDSW = \frac{\sum_{t} IDSW_{t}}{|GT|}.
\end{equation}

A metric that focuses on the entire sequence (how long an object is correctly tracked) is the IDF1 \cite{Ristani16a}, i.e.
\begin{equation}
 IDF1 = \frac{2IDTP}{2IDTP + IDFP + IDFN},
\end{equation}
where IDTP are the correct tracked objects, IDFP are tracked objects that do not match any ground-truth object and IDFN are objects in the ground-truth that are not tracked. The global assignment of tracker hypothesis and ground-truth objects is performed using the suggested truth-to-result match procedure, i.e. the combination that yields the highest IDF1 score is selected.

The Higher Order Tracking Accuracy (HOTA)\cite{Luiten21a},
\begin{equation}
 HOTA_{\alpha} = \sqrt{DetA_{\alpha} \cdot AssA_{\alpha}},
\end{equation}
is an evaluation metric for assessing MOT trackers, where $DetA_{\alpha}$ and $AssA_{\alpha}$ denote the detection and association accuracy at localization threshold $\alpha$. The localization threshold represents the minimum IoU that a detected BB and a ground-truth BB must reach in order to consider the corresponding ground-truth object as detected (i.e., as TP detection). Thus, HOTA combines all three different aspects for MOT tracker evaluation (i.e., localization, detection and association performance) into a single score. To evaluate the different components individually, HOTA can be decomposed into submetrics. In the considered experimental evaluation, only the association performance is relevant (the detected BBs and locations are fixed for all experiments).

To measure the association performance, the authors in \cite{Luiten21a} propose the concept of True Positive Associations (TPAs), False Negative Associations (FNAs) and False Positive Associations (FPAs). TPAs are the set of all correctly tracked TPs (correctly detected objects), while FNAs are the set of TPs that are assigned different trIDs for the same ground truth ID (gtID), and FNs (not detected objects). In other words, FNAs represent the amount of intra-person trID switches. On the contrary, FPAs are the set of TPs with the same trID but different gtIDs (i.e., inter-person trID switches), and FPs (wrongly detected objects). Based on the TPAs, FNAs and FPAs the association recall (AssRe), i.e.
\begin{equation}
 \text{AssRe}_\alpha = \frac{1}{|\text{TP}|} \sum_{c \in \left\{ \text{TP} \right\} } \frac{|\text{TPA}(c)|}{|\text{TPA}(c)| + |\text{FNA}(c)|},
\end{equation}
and association precision (AssPr), i.e.
\begin{equation}
 \text{AssPr}_\alpha = \frac{1}{|\text{TP}|} \sum_{c \in \left\{ \text{TP} \right\} } \frac{|\text{TPA}(c)|}{|\text{TPA}(c)| + |\text{FPA}(c)|},
\end{equation}
can be computed. A combination (Jaccard index) of AssRe and AssPr is the association accuracy (AssA), i.e.
\begin{equation}
 \text{AssA}_\alpha = \frac{\text{AssRe}_\alpha \cdot \text{AssPr}_\alpha}{\text{AssRe}_\alpha + \text{AssPr}_\alpha - \text{AssRe}_\alpha \cdot \text{AssPr}_\alpha}.
\end{equation}
In HOTA, AssA is the main measure for association performance. For all reported results, $\alpha = 0.2$.

The code provided in \cite{Luiten21atrackeval} is used to calculate all evaluation metrics, with the average value (across all sequences) usually being reported.

\section{Experimental Evaluation}
\label{sec:exp_eval}
A comprehensive experimental evaluation is conducted, including a parameter evaluation, a comparison with the state-of-the-art and an ablation study. Using Euclidean distance as similarity metric for association instead of cosine similarity lead to a different scaling of the matching scores. However, the tracking performance does not change significantly. Thus (and due to space constraints), these results are not reported in this work. Interested readers will find these results in the supplementary material together with the complete results, from which all figures and tables are generated. In all tables, the best results are marked in bold.

\subsection{Parameter Evaluation}
\label{sec:para_eval}
In FaceQSORT, the selection of the weighting parameter $\lambda$ (combination of biometric and appearance features) and the association threshold $\theta$ is crucial. Thus, the parameter selection is evaluated in detail along with biometric features extracted by different face recognition models.
\subsubsection{Selection of a fixed $\lambda$}
\label{sec:fixed_lambda}
\begin{figure*}[!t]
\centering
\includegraphics[width=1.0\linewidth]{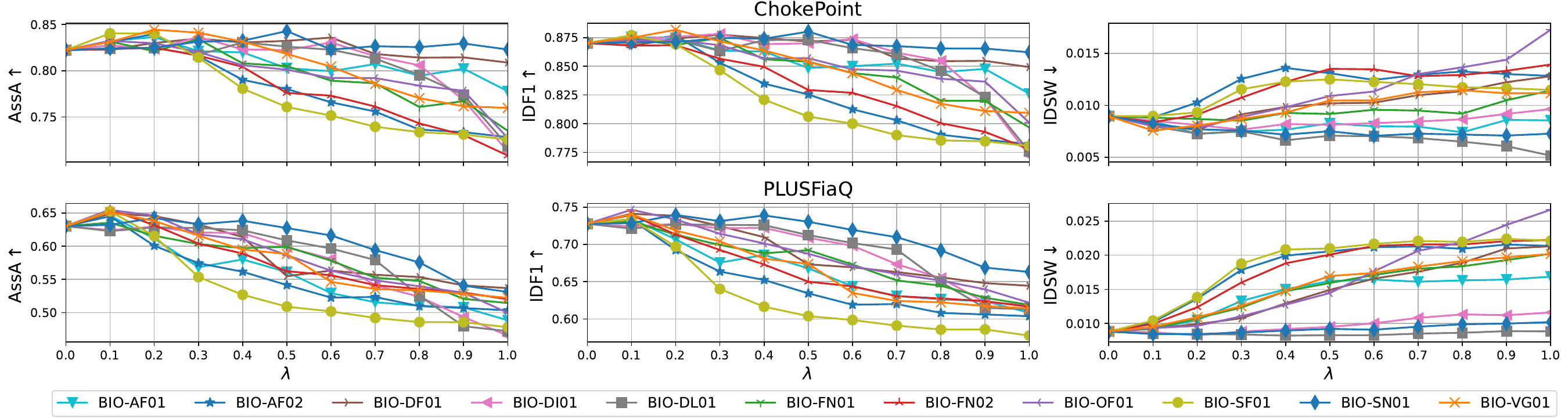}
\caption{The average (across all sequences) AssA and IDF1 scores as well as the normalized amount of IDSW for all evaluated face recognition models and $\lambda$ selections. Recall: for AssA and IDF1, the higher the better, and for IDSW, the lower the better.}
\label{fig:compare_lambdas}
\end{figure*}

\begin{table}
\centering
\caption{PLUSFiaQ: maximum AssA and IDF1 as well as minimum IDSW, with the corresponding $\lambda$ selection.}
\label{tab:compare_lambdas_best_PLUSFiaQ}
\scriptsize
 \begin{tabular}{l || c | c || c | c || c | c}
   & $\lambda$ & AssA$\uparrow$ & $\lambda$ & IDF1$\uparrow$ & $\lambda$ & IDSW$\downarrow$ \\ \hline \hline
BIO-AF01  & 0.1 & 0.6461  & 0.1 & 0.7342  & 0.0 & 0.0088  \\
BIO-AF02  & 0.1 & 0.6451  & 0.1 & 0.7333  & 0.0 & 0.0088  \\ 
BIO-DF01  & 0.1 & 0.6494  & 0.1 & 0.7411  & 0.0 & 0.0088  \\
BIO-DI01  & 0.0 & 0.6303  & 0.0 & 0.7277  & 0.2 & 0.0083  \\
BIO-DL01  & 0.0 & 0.6303  & 0.0 & 0.7277  & \textbf{0.4} & \textbf{0.0082}  \\
BIO-FN01  & 0.1 & 0.6355  & 0.1 & 0.7308  & 0.0 & 0.0088  \\
BIO-FN02  & \textbf{0.1} & \textbf{0.6550}  & 0.1 & 0.7397  & 0.0 & 0.0088  \\
BIO-OF01  & 0.1 & 0.6547  & \textbf{0.1} & \textbf{0.7466}  & 0.0 & 0.0088  \\
BIO-SF01  & 0.1 & 0.6527  & 0.1 & 0.7338  & 0.0 & 0.0088  \\
BIO-SN01  & 0.2 & 0.6425  & 0.2 & 0.7396  & 0.2 & 0.0084  \\
BIO-VG01  & 0.1 & 0.6515  & 0.1 & 0.7405  & 0.0 & 0.0088  \\
\hline \hline
 \end{tabular}
\end{table}

\begin{table}
\centering
\caption{ChokePoint: maximum AssA and IDF1 as well as minimum IDSW, with the corresponding $\lambda$ selection.}
\label{tab:compare_lambdas_best_ChokePoint}
\scriptsize
 \begin{tabular}{l || c | c || c | c || c | c}
   & $\lambda$ & AssA$\uparrow$ & $\lambda$ & IDF1$\uparrow$ & $\lambda$ & IDSW$\downarrow$ \\ \hline \hline
BIO-AF01  & 0.2 & 0.8364  & 0.1 & 0.8754  & 0.3 & 0.0068  \\
BIO-AF02  & 0.2 & 0.8396  & 0.1 & 0.8777  & 0.0 & 0.0076  \\
BIO-DF01  & 0.6 & 0.8359  & 0.3 & 0.8755  & 0.0 & 0.0076  \\
BIO-DI01  & 0.6 & 0.8306  & 0.2 & 0.8756  & 0.0 & 0.0076  \\
BIO-DL01  & 0.5 & 0.8353  & 0.1 & 0.8752  & \textbf{1.0} & \textbf{0.0052}  \\
BIO-FN01  & 0.3 & 0.8306  & 0.0 & 0.8724  & 0.0 & 0.0076  \\
BIO-FN02  & 0.1 & 0.8355  & 0.1 & 0.8773  & 0.1 & 0.0075  \\
BIO-OF01  & 0.2 & 0.8298  & 0.2 & 0.8728  & 0.0 & 0.0076  \\
BIO-SF01  & 0.1 & 0.8404  & 0.1 & 0.8783  & 0.0 & 0.0076  \\
BIO-SN01  & 0.4 & 0.8325  & 0.4 & 0.8740  & 0.5 & 0.0069  \\
BIO-VG01  & \textbf{0.3} & \textbf{0.8497}  & \textbf{0.2} & \textbf{0.8819}  & 0.0 & 0.0076  \\
\hline \hline
 \end{tabular}
\end{table}

\begin{figure}[!t]
\centering
\includegraphics[width=1.0\linewidth]{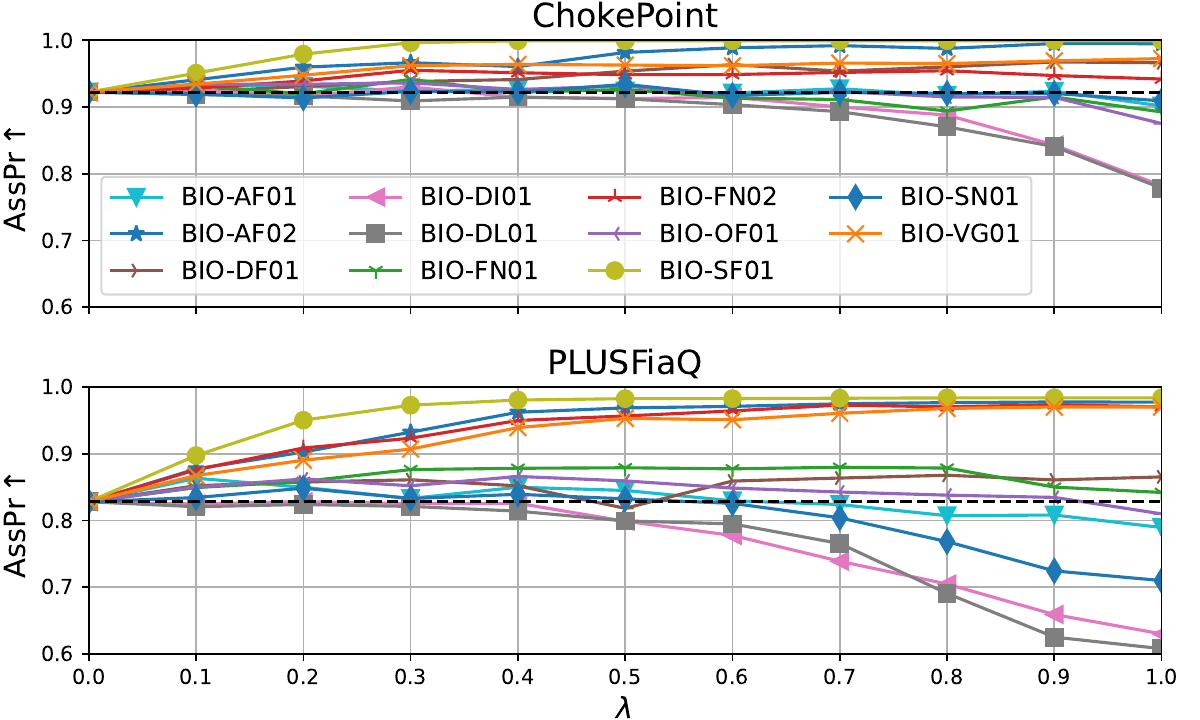}
\caption{Average AssPr (across all evaluated sequences) for different $\lambda$ selections.}
\label{fig:AssPr_fixed_lambda}
\end{figure}

Figure \ref{fig:compare_lambdas} shows the average (across all sequences) AssA and IDF1 scores as well as the normalized amount of IDSW for all evaluated face recognition models and different $\lambda$ selections. In general, the higher the AssA or IDF1 score or the lower the number of IDSWs, the better the tracking performance. The scores achieved when setting $\lambda$ to 1.0 can be considered as the baseline. In this setting, only biometric features are taken into account. When looking at AssA and IDF1, a general increase in scores can be observed with increasing influence of the appearance feature (i.e., $\lambda \rightarrow 0$). The number of IDSWs, on the other hand, decreases. Thus, the tracking performance increases when $\lambda$ decreases (i.e., with increasing influence of the appearance feature).

The best values achieved (i.e., max. AssA and IDF1 as well as min. IDSW) are shown together with the corresponding $\lambda$ values in tables \ref{tab:compare_lambdas_best_PLUSFiaQ} and \ref{tab:compare_lambdas_best_ChokePoint}. In general, significantly higher results are achieved with the ChokePoint dataset (i.e., the PLUSFiaQ dataset is more challenging). With the PLUSFiaQ dataset, the best results (AssA and IDF1) are achieved at $\lambda = 0.1$ (except for BIO-DI01, BIO-DL01 and BIO-SN01). In contrast, the best results are achieved with the Chokepoint dataset having $\lambda$ values up to 0.6. One reason for this could be that with the ChokePoint dataset, people look towards the camera when they approach the portal, which facilitates the usage of biometric features. The number of IDSWs is usually lowest at $\lambda = 0.0$ (i.e., when only appearance features are used). However, it can be stated that the combination of biometric and appearance features improves tracking performance for both datasets.

In the considered scenario, inter-person trID switches (i.e., when the trID is switched to an existing trID of a different person) are more critical than intra-person trID switches (i.e., when a new trID is assigned). As already described, the AssPr score reflects the FPAs (i.e., inter-person trID switches) and is illustrated in Figure \ref{fig:AssPr_fixed_lambda}. If only biometric features are used for association (i.e., $\lambda = 1.0$), inter-person trID switches are expected to be limited, as the extracted biometric features should be distinct for different identities. For both analyzed datasets, this can basically be observed for 4/11 evaluated face recognition models (i.e., BIO-SF01, BIO-AF02, BIO-FN02 and BIO-VG01), with BIO-SF01 features showing the most consistent and best results. However, the AssPr decreases as $\lambda$ decreases. This is reasonable as the association in reducing $\lambda$ is more and more based on generic appearance features and the appearance of two different identities may be similar, while the biometric features are distinct. Considering that the best AssA for these 4 face recognition models is reached at lower $\lambda$ values (i.e., 0.1 to 0.3 for the ChokePoint and 0.1 for the PLUSFiaQ dataset), the higher AssA is accompanied by a decrease in AssPr (i.e., an increase in inter-person trID switches (FPAs)). In contrast, the AssPr for BIO-DL01 and BIO-DI01 is much lower for $\lambda = 1.0$ than compared to the appearance features baseline (Figure \ref{fig:AssPr_fixed_lambda} dotted black line). This would indicate a weak distinctiveness of these biometric features for the considered scenario.

\subsubsection{Selection of a dynamic $\lambda$ based on Face Quality}
\begin{figure}[!t]
\centering
\includegraphics[width=0.98\linewidth]{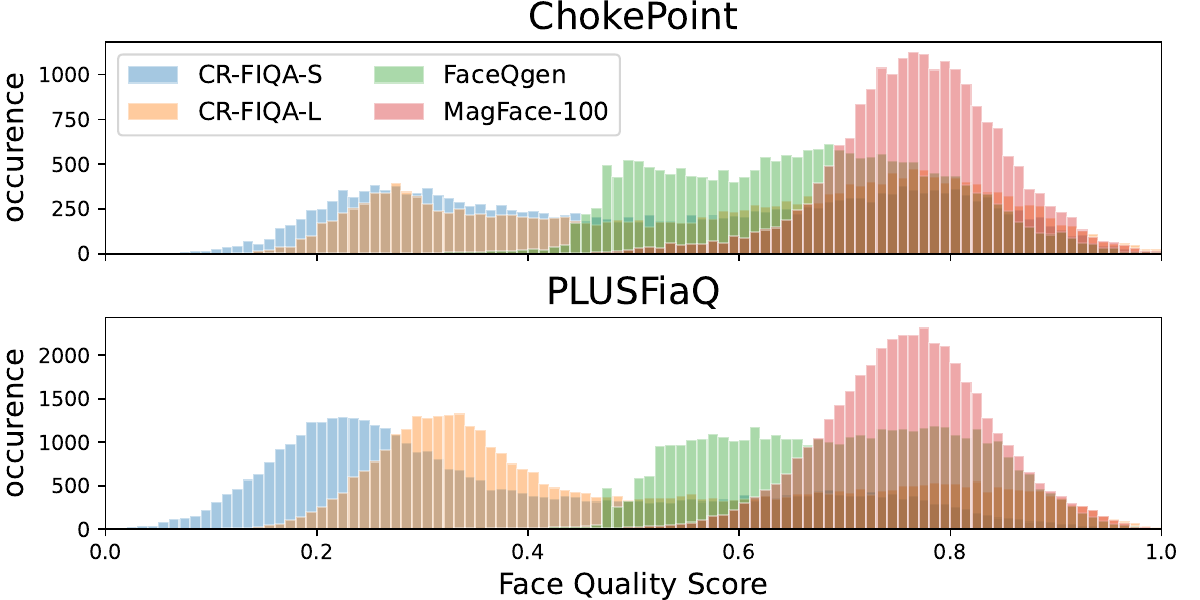}
\caption{Quality score distribution.}
\label{fig:quality_socre_dist}
\end{figure}
\noindent By setting the parameter $\lambda$ to a fixed value (e.g., $\lambda = 0.1$), the weighting between biometric and appearance features is constant for each image and each detected face, regardless of the biometric information present in the extracted face patch. For this reason, a more natural approach would be for $\lambda$ to reflect the available biometric information (i.e., $\lambda=1$ when full biometric information is available and $\lambda=0$ when no biometric information is available). The biometric information present in a face sample can be quantified by a face quality score \cite{Boutros23a,Meng21a,Hernandez21a}. For this reason, another option would be to set $\lambda$ dynamically according to a face quality score. This turns $\lambda$ into an $|\Phi| \times |\Psi|$ matrix with constant columns corresponding to the face quality scores of the detected faces $\Psi$. The multiplications $\lambda C_{bio}$ and $(1- \lambda) C_{app}$ are then element wise multiplications.

Three different face quality models are exploited, i.e. CR-FIQA \cite{Boutros23a}, MagFace \cite{Meng21a} and FaceQgen \cite{Hernandez21a}. In case of the CR-FIQA two different backbone model sizes are evaluated, i.e. CR-FIQA-S and CR-FIQA-L. To obtain a face quality score, the detected face patch is fed into the respective model. The distribution of quality scores achieved for all detected faces (across all sequences) is illustrated in Figure \ref{fig:quality_socre_dist}. Quality values that are not restricted to $\mathbb{R}_{[0,1]}$ are normalized by dividing by the maximum observed value.

\begin{figure*}[!t]
\centering
\includegraphics[width=0.98\linewidth]{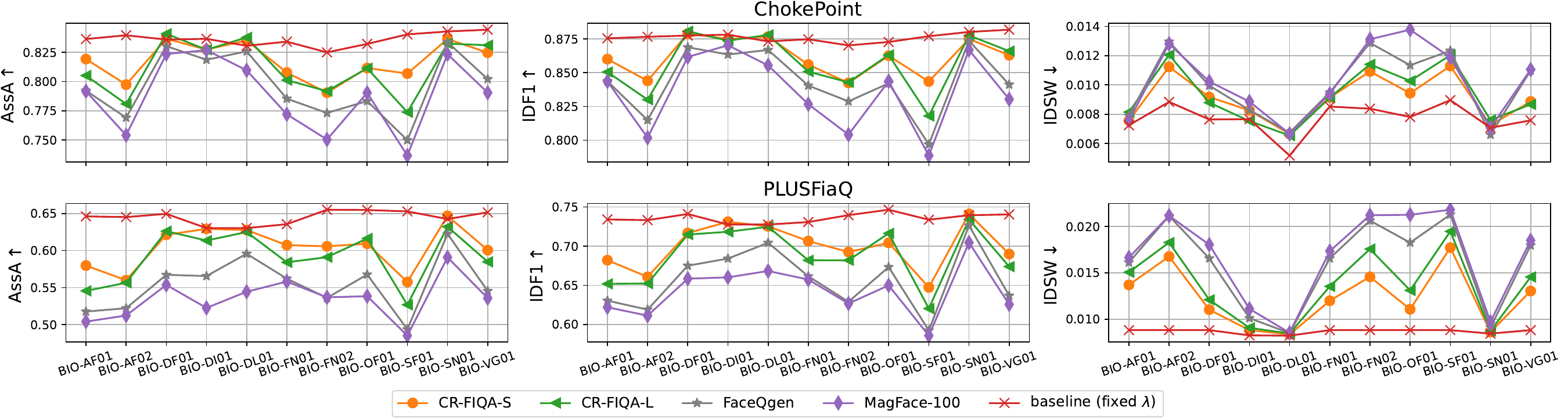}
\caption{Achieved AssA, IDF1 and IDSW with $\lambda$ dynamically set based on normalized (by the maximum value) face quality scores compared to the best achieved AssA, IDF1 and IDSW based on a fixed $\lambda$ value. Recall: for AssA and IDF1, the higher the better, and for IDSW, the lower the better.}
\label{fig:face_quality_vs_fixed_lambda}
\end{figure*}

In Figure \ref{fig:face_quality_vs_fixed_lambda}, the results are compared between a fixed (best results table \ref{tab:compare_lambdas_best_PLUSFiaQ} and \ref{tab:compare_lambdas_best_ChokePoint}) and a dynamically set (based on face quality scores) $\lambda$. It can be seen that tracking performance is generally lower when $\lambda$ is dynamically set based on face quality scores. Only with CR-FIQA the performance for some face recognition models is as high as when $\lambda$ is set to a fixed value. In general, the performance based on CR-FIQA is consistently better than the other two face quality models evaluated. However, the reasonable approach of dynamically setting $\lambda$ based on available biometric information (face quality scores) is not working. One reason for this could be that face quality metrics are designed to predict face recognition performance, which does not directly translate to multi-face tracking.

\subsubsection{Grid search dynamic $\lambda$}
\begin{figure}[!t]
\centering
\includegraphics[width=0.98\linewidth]{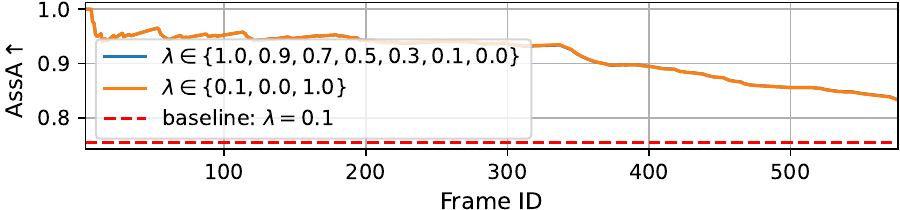}
\caption{Highest AssA achieved per frame in the grid search (sequence 4).}
\label{fig:grid_search_AssA}
\end{figure}
\noindent To assess whether a higher AssA can be achieved with a dynamic $\lambda$ (which is different for each detected face) than with a fixed $\lambda$, a grid search is performed. In a grid search, all possible combinations of $\lambda$ are evaluated. For example, if 5 faces are detected per frame and 11 different $\lambda$ values are allowed (i.e., $\lambda \in [0.0,0.1,0.2,0.3,0.4,0.5,0.6,0.7,0.8,0.9,1.0]$), this would result in $11^5$ different combinations for the frame under consideration. Since the association in the current frame depends on the past frames, the different combinations must be considered across all frames, i.e. $\prod_{i=1}^{N} 11^{m^i}$ combinations, where $N$ is the number of frames and $m^i$ are the detected faces for the $i$-th frame. This is simply not computable. Thus, only a local grid search (between two consecutive frames) is performed. More precisely, the AssA is calculated after each $\lambda$ combination for the current frame. The combination that achieves the highest AssA is selected. If several combinations achieve the highest AssA, the combination that achieves the highest AssA first is selected.

\begin{figure}[!t]
\centering
\includegraphics[width=0.88\linewidth]{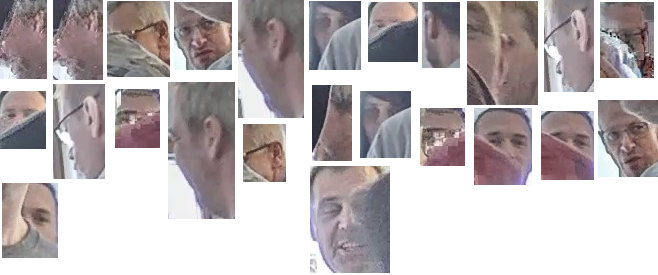}
\caption{Face patches that were assigned a $\lambda$ of 1.0 during the grid search with sequence 4. Since only biometric features are used for $\lambda = 1.0$, it is expected that the corresponding face patches contain rich biometric information, which is not the case.}
\label{fig:grid_search_example_faces}
\end{figure}

Since the evaluation of 11 different $\lambda$ values is still very time-consuming, a smaller number is considered. In addition, the grid search is only performed with the best face recognition model (in terms of AssPr), BIO-SF01, and only for the two sequences with the fewest frames (sequence ID 3 and 4). In Figure \ref{fig:grid_search_AssA}, the highest AssA achieved for each frame of sequence 4 is illustrated. It can be seen that the AssA at frame 576 (complete sequence) is about 0.09 higher than the baseline (fixed $\lambda$) and exactly the same results are obtained regardless of whether $\lambda$ is selected from \{1.0,0.9,0.7,0.5,0.3,0.1,0.0\} or from \{0.1,0.0,1.0\}.

Taking into account possible $\lambda$ values of \{1.0,0.9,0.7,0.5,0.3,0.1,0.0\}, a $\lambda$ of 1.0 is selected for 2199/2239 detected faces in sequence 4. The $\lambda$ values for the remaining 40 detected faces are distributed as follows: 5, 22, 7, 4, 2 for the $\lambda$ values 0.0, 0.1, 0.3, 0.5 and 0.7, respectively. The vast majority of $\lambda$ values are set to 1.0 because the grid search conducted always selects the first $\lambda$ combination that achieves the highest AssA value, and the first combination tested is that all $\lambda$ values equal 1.0. In the version where $\lambda$ can be chosen from only three values (i.e., \{0.1,0.0,1.0\}), the first combination tested is that all $\lambda$ values are equal to 0.1, and the last combination is that all $\lambda$ values are equal to 1.0 (i.e., a $\lambda$ of 1.0 is only chosen if 0.1 and 0.0 do not achieve an equally high AssA). In this case, a $\lambda$ of 0.1 is selected for 2206/2239 detected faces. Compared to the baseline (i.e., fixed $\lambda$ of 0.1 for all detected faces), a change of $\lambda$ for only 33 detected faces yields the performance gain. In particular, for 24 detected faces, a $\lambda$ value of 1.0 is selected. A $\lambda$ value of 1.0 means that only biometric features are used. Therefore, it is expected that the corresponding face patches contain adequate biometric information. The 24 face patches where $\lambda$ is set to 1.0 are illustrated in Figure \ref{fig:grid_search_example_faces}. As can be seen, the biometric information is very limited.

The results for sequence 3 are similar. These observations indicate that linking $\lambda$ to the available biometric information does not lead to better performance.

\subsubsection{Matching score distribution}
\begin{table}
\centering
\caption{Presents the overlap (IoU) between the genuine and imposter scores as well as their mean ($\mu$) and standard deviation ($\sigma$). Feature models: (1) BIO-AF01, (2) BIO-AF02, (3) BIO-DF01, (4) BIO-DI01, (5) BIO-DL01, (6) BIO-FN01, (7) BIO-FN02, (8) BIO-OF01, (9) BIO-SF01, (10) BIO-SN01, (11) BIO-VG01 and (12) Appearance Features.}
\label{tab:IvG_dist}
\scriptsize
 \begin{tabular}{m{1.4em} | c | c | c || c | c | c }
 \multicolumn{1}{c}{} & \multicolumn{3}{c}{PLUSFiaQ} & \multicolumn{3}{c}{ChokePoint}\\
 \multicolumn{1}{c}{} & \multicolumn{1}{c}{} & \multicolumn{1}{c}{genuine} & \multicolumn{1}{c}{imposter} & \multicolumn{1}{c}{} & \multicolumn{1}{c}{genuine} & \multicolumn{1}{c}{imposter} \\
 & IoU & $\mu/\sigma$ & $\mu/\sigma$ & IoU & $\mu/\sigma$ & $\mu/\sigma$\\ \hline \hline
(1) & 0.26 & 0.14/0.14 & 0.52/0.31 & 0.20 & 0.15/0.14 & 0.57/0.29 \\
(2) & 0.08 & 0.21/0.15 & 0.77/0.18 & 0.05 & 0.23/0.14 & 0.82/0.16 \\
(3) & 0.14 & 0.10/0.07 & 0.25/0.08 & 0.12 & 0.10/0.08 & 0.26/0.08 \\
(4) & 0.20 & 0.02/0.03 & 0.09/0.06 & 0.17 & 0.03/0.04 & 0.10/0.06 \\
(5) & 0.07 & 0.02/0.01 & 0.07/0.02 & 0.08 & 0.02/0.01 & 0.08/0.03 \\
(6) & 0.18 & 0.12/0.12 & 0.51/0.25 & 0.26 & 0.13/0.13 & 0.46/0.27 \\
(7) & 0.07 & 0.16/0.14 & 0.76/0.20 & 0.09 & 0.19/0.15 & 0.73/0.22 \\
(8) & 0.18 & 0.10/0.08 & 0.31/0.14 & 0.18 & 0.11/0.09 & 0.35/0.17 \\
(9) & \textbf{0.04} & 0.27/0.17 & 0.87/0.13 & \textbf{0.04} & 0.29/0.17 & 0.90/0.12 \\
(10) & 0.18 & 0.02/0.03 & 0.09/0.06 & 0.19 & 0.05/0.07 & 0.25/0.16 \\
(11) & 0.07 & 0.15/0.11 & 0.60/0.15 & 0.08 & 0.20/0.12 & 0.67/0.17 \\
\hline
(12) & 0.16 & 0.03/0.03 & 0.28/0.19 & 0.18 & 0.03/0.04 & 0.26/0.18 \\
\hline \hline
\end{tabular}
\end{table}

\noindent A multi-face tracker can be compared with a biometric system (e.g., a set of query images (detected faces) are matched against a database of enrolled identities (active tracks)). Matching is based on a similarity score, and in a perfect system, the genuine scores do not overlap with the impostor scores. In the context of FaceQSORT, the genuine scores represent the cosine similarity that result when a detected face is compared with the corresponding stored faces (active tracks). The imposter scores, on the other hand, represent the cosine similarities that are achieved when a detected face is compared with all other stored faces (tracks). Table \ref{tab:IvG_dist} shows the mean and the standard deviation of the genuine and the imposter score distribution, as well as their overlap (IoU). Considering the IoU, the most distinct face recognition model (lowest IoU) is BIO-SF01. Other face recognition models with a low IoU are BIO-FN02, BIO-VG01, BIO-AF02 and BIO-DL01. These face recognition models (except BIO-DL01) are also the models achieving the highest AssPr (see Figure \ref{fig:AssPr_fixed_lambda}). However, based on the IoU results, it is surprising that such low AssPr values are obtained with the Dlib face recognition model. A reason for the low AssPr values could be that, with a mean value of 0.07, the imposter scores are well below the threshold $\theta = 0.2$ (applied to reject uncertain (imposter) matches). Thus, a closer look is taken at the selection of the threshold value $\theta$.

\subsubsection{Selection of the threshold $\theta$}
\begin{table*}
\centering
\caption{Best achieved AssA, IDF1 and IDSW for different selections of $\theta$ and $\lambda$ combinations. Face recognition models: (1) BIO-AF01, (2) BIO-AF02, (3) BIO-DF01, (4) BIO-DI01, (5) BIO-DL01, (6) BIO-FN01, (7) BIO-FN02, (8) BIO-OF01, (9) BIO-SF01, (10) BIO-SF01 and (11) BIO-VG01.}
\label{tab:compare_theta_lambda_cosine}
\scriptsize
 \begin{tabular}{l || c | c || c | c || c | c ||| c | c || c | c || c | c}
  \multicolumn{1}{c}{} & \multicolumn{6}{c}{PLUSFiaQ} & \multicolumn{6}{c}{ChokePoint}\\
   & $\theta$/$\lambda$ & AssA$\uparrow$  & $\theta$/$\lambda$ & IDF1$\uparrow$  & $\theta$/$\lambda$ & IDSW$\downarrow$
   & $\theta$/$\lambda$ & AssA$\uparrow$  & $\theta$/$\lambda$ & IDF1$\uparrow$  & $\theta$/$\lambda$ & IDSW$\downarrow$\\ \hline \hline
(1) & 0.2/0.1 & 0.6461& 0.2/0.1 & 0.7342& 0.35/0.1 & 0.0080& 0.2/0.2 & 0.8364& 0.2/0.1 & 0.8754& 0.56/0.1 & 0.0054 \\
(2) & 0.3/0.2 & 0.6572& 0.3/0.2 & 0.7433& 0.4/0.2 & 0.0078& 0.4/0.5 & 0.8611& 0.4/0.5 & 0.8928& 0.6/0.6 & 0.0047 \\
(3) & 0.2/0.1 & 0.6494& 0.2/0.1 & 0.7411& 0.3/0.1 & 0.0078& 0.25/0.4 & 0.8446& 0.25/0.4 & 0.8840& 0.3/0.2 & 0.0053 \\
(4) & 0.175/0.1 & 0.6399& 0.175/0.1 & 0.7349& 0.2/0.2 & 0.0083& 0.15/0.4 & 0.8376& 0.15/0.4 & 0.8778& 0.2/0.3 & 0.0077 \\
(5) & 0.2/0.2 & 0.6283& 0.2/0.2 & 0.7270& 0.2/0.4 & 0.0082& 0.2/0.5 & 0.8353& 0.2/0.1 & 0.8752& 0.2/0.9 & 0.0059 \\
(6) & 0.21/0.1 & 0.6357& 0.21/0.1 & 0.7337& 0.35/0.1 & 0.0079& 0.21/0.1 & 0.8314& 0.21/0.1 & 0.8759& 0.56/0.5 & 0.0051 \\
(7) & \textbf{0.3/0.3} & \textbf{0.6693}& 0.3/0.3 & 0.7520& 0.4/0.2 & 0.0078& 0.3/0.3 & 0.8493& 0.4/0.4 & 0.8812& 0.7/0.2 & 0.0049 \\
(8) & 0.25/0.3 & 0.6589& 0.25/0.3 & 0.7505& 0.3/0.2 & 0.0077& 0.2/0.2 & 0.8298& 0.2/0.2 & 0.8728& 0.35/0.4 & 0.0052 \\
(9) & 0.4/0.4 & 0.6663& \textbf{0.4/0.3} & \textbf{0.7559}& \textbf{0.6/0.5} & \textbf{0.0072}& \textbf{0.5/0.5} & \textbf{0.8879}& \textbf{0.5/0.5} & \textbf{0.9075}& \textbf{0.8/0.2} & \textbf{0.0046} \\
(10) & 0.2/0.2 & 0.6425& 0.2/0.2 & 0.7396& 0.25/0.3 & 0.0077& 0.25/0.7 & 0.8457& 0.25/0.3 & 0.8811& 0.35/0.6 & 0.0050 \\
(11) & 0.2/0.1 & 0.6515& 0.2/0.1 & 0.7405& 0.3/0.1 & 0.0079& 0.3/0.6 & 0.8581& 0.3/0.6 & 0.8875& 0.8/0.8 & 0.0048 \\ \hline
$\varnothing$ & 0.24/0.19 & 0.6496 & 0.24/0.18 & 0.7412 & 0.33/0.22 & 0.0078 & 0.27/0.40 & 0.8470 & 0.28/0.33 & 0.8828 & 0.49/0.44 & 0.0054\\
\hline \hline
 \end{tabular}
\end{table*}

\begin{figure}[!t]
\centering
\includegraphics[width=0.98\linewidth]{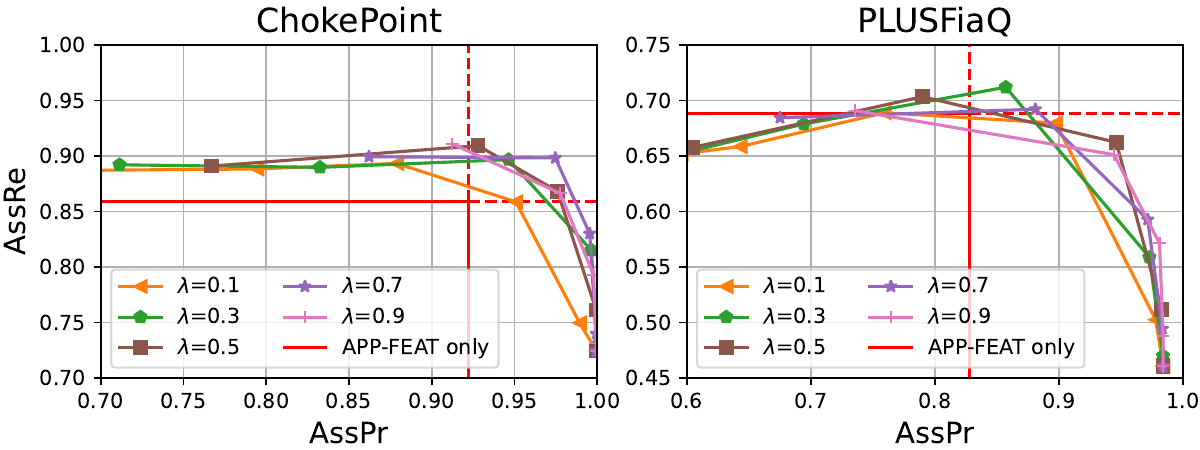}
\caption{AssRe over AssPr for different $\theta$ and $\lambda$ combinations and BIO-SF01 features. The $\theta$ values increase from left to right and the red lines represent the baseline (AssRe and AssPr of the best achieved AssA, if only appearance features are used).}
\label{fig:AssRevsAssPr_cosine}
\end{figure}

\noindent For both datasets, the genuine and impostor scores obtained on the basis of BIO-DL01 features are close to zero (see Table \ref{tab:IvG_dist}). Thus, with a threshold $\theta = 0.2$ possible imposter matches (inter-person trID switches) are not rejected. This explains the low AssPr value when only BIO-DL01 features are used for the association (see `1) Selection of a fixed $\lambda$' of subsection \ref{sec:para_eval}), although the genuine and imposter distributions are relatively distinct (an IoU value of 0.07 and 0.08). If $\theta$ is set to 0.025, an AssPr of 0.9831 and 0.9938 is achieved for the PLUSFiaQ and ChokePoint datasets, respectively. This shows that, in addition to $\lambda$, $\theta$ is an important parameter as well. In Table \ref{tab:compare_theta_lambda_cosine}, the best achieved AssA, IDF1 and IDSW values for different $\theta$ and $\lambda$ combinations are shown. It can be seen that the initial choice of $\theta = 0.2$ was already pretty good for several face recognition models. Overall, the best results are achieved with BIO-SF01 features, which also provides the most distinctive genuine and imposter score distributions (IoU of 0.04). Again, the $\lambda$ values that yield the best results are significantly higher for the ChokePoint dataset (probably fewer occlusions and lateral faces benefit the usage of biometric features). Although very low $\theta$ values (e.g. 0.025, 0.05, etc.) are tested, a combination of BIO-DL01 and appearance features performs worst overall.

When considering the threshold $\theta$ the expectation is that the AssPr is very high for low $\theta$ values (reject most imposter matches) and it decreases when $\theta$ increases. However, a low value of $\theta$ can lead to a lower AssRe (more intra-person trID switches) since also genuine matches are rejected. In Figure \ref{fig:AssRevsAssPr_cosine} the trade-off between AssRe and AssPr for the best face recognition model BIO-SF01 is shown for different $\lambda$ and $\theta$ combinations. The $\theta$ values decrease from left to right (i.e., 0.9, 0.8, 0.7, 0.6, 0.5, 0.4, 0.3, 0.2, 0.1, 0.05 and 0.025) and the red lines show the AssRe and AssPr for the best AssA achieved with appearance features only (baseline). A good trade-off (relatively constant high AssPr while reducing intra-person trID switches (AssRe)) when increasing the threshold $\theta$ can be achieved for higher $\lambda$ values (higher contribution of biometric features).

\subsection{Tracker Comparison}
\label{sec:exp_results}
\begin{table*}
\centering
\caption{Compare FaceQSORT with state-of-the-art tracker, where \sout{MC} and \sout{IoU} denote FaceQSORT without matching cascade (MC) and/or IoU fallback matching. StrongSORT and DeepSORT are applied either with biometric (BIO-SF01) or with appearance features.}
\label{tab:compare_tracker}
\scriptsize
 \begin{tabular}{l | c | c | c | c || c | c | c | c || c | c | c | c}
 \multicolumn{1}{c}{} & \multicolumn{4}{c}{ PLUSFiaQ} & \multicolumn{4}{c}{ChokePoint} & \multicolumn{4}{c}{MusicVideo} \\
  & AssA$\uparrow$ & HOTA$\uparrow$ & IDF1$\uparrow$ & IDSW$\downarrow$ & AssA$\uparrow$ & HOTA$\uparrow$ & IDF1$\uparrow$ & IDSW$\downarrow$ & AssA$\uparrow$ & HOTA$\uparrow$ & IDF1$\uparrow$ & IDSW$\downarrow$ \\ \hline \hline
FaceQSORT & \textbf{0.6527}& \textbf{0.7486} & \textbf{0.7338}& \textbf{0.0104}& 0.8404& 0.8671& \textbf{0.8771}& 0.0090& 0.0425& 0.1821& 0.0918& 0.0343 \\
StrongSORT$_{APP}$\cite{Du23a} & 0.4825& 0.6303& 0.6172& 0.0177& 0.6626& 0.7657& 0.7729& 0.0158& 0.0407& 0.1720& 0.0841& 0.0363 \\
StrongSORT$_{BIO}$\cite{Du23a} & 0.4780& 0.6256& 0.5678& 0.0198& 0.7300& 0.8080& 0.7750& 0.0099& 0.0305& 0.1486& 0.0637& 0.0328 \\
DeepSORT$_{APP}$\cite{Wojke17a} & 0.5556& 0.6805& 0.6769& 0.0094& 0.7925& 0.8418& 0.8453& 0.0075& 0.0429& 0.1769& 0.0914& 0.0287 \\
DeepSORT$_{BIO}$\cite{Wojke17a} & 0.4700& 0.6201& 0.5570& 0.0191& 0.7278& 0.8063& 0.7740& 0.0112& 0.0306& 0.1488& 0.0634& 0.0325 \\
DeepOCSORT\cite{Maggiolino23a} & 0.4816& 0.6114& 0.5819& 0.0172& 0.7841& 0.8244& 0.8225& 0.0069& 0.0354& 0.1585& 0.0774& 0.0318 \\
OCSORT\cite{Cao23a} & 0.5030& 0.6292& 0.6006& 0.0143& 0.8280& 0.8487& 0.8502& 0.0051& 0.0362& 0.1598& 0.0764& 0.0310 \\
BoTSORT\cite{Aharon22a} & 0.4206& 0.5257& 0.4880& 0.0265& 0.7092& 0.6959& 0.6760& 0.0139& 0.0301& 0.1397& 0.0614& 0.0366 \\
SORT\cite{Bewley16a} & 0.5250& 0.6558& 0.6006& 0.0148& \textbf{0.8583}& \textbf{0.8804}& 0.8612& 0.0051& 0.0336& 0.1543& 0.0651& 0.0312 \\
OMTMCM\cite{Weng23a} & 0.2039& 0.3763& 0.3352& 0.1671& 0.2430& 0.4483& 0.4140& 0.1335& \textbf{0.1076}& \textbf{0.2841}& \textbf{0.2383}& 0.0944 \\
UMA\cite{Yin20a} & 0.1828& 0.2475& 0.2640& 0.0437& 0.5481& 0.6929& 0.6788& 0.0112& 0.0490& 0.1930& 0.0988& \textbf{0.0268} \\
ByteTrack\cite{Zhang22a} & 0.4846& 0.6163& 0.5831& 0.0145& 0.8320& 0.8509& 0.8520& \textbf{0.0046}& 0.0349& 0.1576& 0.0740& 0.0287 \\
\hline
FaceQSORT\sout{IoU} & 0.5764& 0.7033& 0.6760& 0.0187& 0.6933& 0.7863& 0.7671& 0.0283& 0.0363& 0.1676& 0.0830& 0.0494 \\
FaceQSORT\sout{MC} & 0.5940& 0.7147& 0.7025& 0.0169& 0.8274& 0.8603& 0.8700& 0.0107& 0.0405& 0.1776& 0.0875& 0.0400 \\
FaceQSORT\sout{MC}\sout{IoU} & 0.5589& 0.6924& 0.6666& 0.0292& 0.6697& 0.7721& 0.7521& 0.0392& 0.0359& 0.1664& 0.0812& 0.0618 \\
\hline \hline
\end{tabular}
\end{table*}
\noindent In table \ref{tab:compare_tracker} FaceQSORT is compared with different state-of-the-art trackers. The FaceQSORT results are obtained with BIO-SF01 features, $\lambda = 0.1$ and $\theta = 0.2$. BIO-SF01 features are selected because these features perform best in the considered scenario. StrongSORT and DeepSORT are also evaluated with BIO-SF01 or appearance features (i.e., StrongSORT$_{APP}$, StrongSORT$_{BIO}$, DeepSORT$_{APP}$ and DeepSORT$_{APP}$). If possible, similar parameters (e.g., $\theta$, $N_{max}=100$, etc.) are set identically for all trackers evaluated.

Implementations are available for SORT\footnote{https://github.com/abewley/sort}, DeepSORT\footnote{https://github.com/nwojke/deep\_sort}, StrongSORT\footnote{https://github.com/dyhBUPT/StrongSORT} and UMA\footnote{https://github.com/yinjunbo/UMA-MOT}. A multi-face tracking version of DeepOCSORT \cite{Maggiolino23a}, OCSORT \cite{Cao23a}, BoTSORT \cite{Aharon22a} and ByteTrack \cite{Zhang22a} with SphereFace features and yolov8 face detection is available in\footnote{https://github.com/yjwong1999/OpenVINO-Face-Tracking-using-YOLOv8-and-DeepSORT} \cite{Wong23a}. No implementation is available for the OMTMCM multi-face tracker. For this reason, OMTMCM is carefully implemented according to the description in \cite{Weng23a}. To verify the implementation, the results for the MusicVideo dataset \cite{Zhang16b} are also presented in table \ref{tab:compare_tracker}. The MusicVideo dataset consists of 8 music videos (i.e., Apink, BrunoMars, Darling, GirlsAloud, HelloBubble, PussycatDolls, Tara and Westlife) from YouTube, is publicly available\footnote{https://sites.google.com/site/shunzhang876/eccv16-face-tracking} and was used in \cite{Weng23a} to evaluate the proposed OMTMCM multi-face tracker. Since music videos are unconstrained videos (with different scenes, moving camera, etc.), they do not correspond to the considered scenario in which people move towards a gate for which FaceQSORT is designed.

For the MusicVideo dataset, OMTMCM performed best. However, in the considered scenario where people are moving towards a gate (PLUSFiaQ), FaceQSORT clearly outperformed the other evaluated state-of-the-art trackers. With the ChokePoint dataset, SORT achieved the highest AssA and HOTA scores. The high performance of SORT with the ChokePoint dataset can be explained by the fact that in the ChokePoint dataset, people walk through the portal with constant (linear) motion. However, FaceQSORT achieved a similarly high AssA and HOTA score and a higher IDF1 score.

\subsection{Ablation Study}
  \noindent The main components of FaceQSORT are the combination of biometric and appearance features, the matching cascade (MC) and IoU fallback matching. Running FaceQSORT with either biometric (i.e., $\lambda = 1.0$) or appearance features (i.e., $\lambda = 0.0$) has already been evaluated in `1) Selection of a fixed $\lambda$' of subsection \ref{sec:para_eval}. Applying the MC carries the risk that not the best matches are found. For example, it could happen that the best match (lowest cost) for a detected face is with a track that could not be matched in the last frames (e.g., a person returning from an occlusion), but the detected face is matched with a track that is considered in the MC before. The achieved tracking performance scores for FaceQSORT (with BIO-SF01 features, $\lambda = 0.1$ and $\theta = 0.2$)  without the matching cascade (\sout{MC}) are reported in table \ref{tab:compare_tracker}. Compared to results when using the matching cascade (table \ref{tab:compare_tracker} first row) a significant performance drop can be observed.

  IoU fallback matching is applied to all unmatched detections after the matching cascade, including matching with tentative tracks (only confirmed tracks are considered in the matching cascade). To achieve the reported tracking performance of FaceQSORT (table \ref{tab:compare_tracker} first row), IoU matching is performed for 3.91\%, 4.9\% and 8.42\% of all detected faces of the PLUSFiaQ, ChokePoint and MusicVideo dataset respectively. When disabling IoU fallback matching (\sout{IoU}) the performance decreases significantly (see table \ref{tab:compare_tracker}). The performance achieved when FaceQSORT is applied without matching cascade and IoU fallback matching is reported in table \ref{tab:compare_tracker} (last row).

\section{Conclusion}
\label{sec:con}
\noindent In this work, a new multi-face tracking method, FaceQSORT, is proposed. To mitigate tracking challenges, two different features (i.e., biometric and appearance features) extracted from the same image (face) patch are combined. It is shown that a combination of the two features is beneficial for multi-face tracking compared to using only one feature type. In the considered scenario, when people move towards a gate (PLUSFiaQ) FaceQSORT clearly outperformed the evaluated state-of-the-art tracker. To get a deeper insight into the proposed method, a comprehensive experimental evaluation and an ablation study are conducted. It is shown that the selection of the face recognition model, as well as the resulting distribution of genuine and imposter scores are crucial. In general, selecting a low similarity threshold $\theta$ and a high parameter $\lambda$ lead to a high AssPr.

For future work, an adaptive $\lambda$ could be evaluated based on the detected faces, i.e. in crowded scenes with probably more partially occluded and lateral faces, $\lambda$ could automatically decrease. Furthermore, the general similarity threshold $\theta$ could be split into two thresholds, one for biometric and one for appearance features, to better account for the different distributions of imposter and genuine scores. A different idea would be to perform association based on biometric features first and apply appearance based association as fallback.
\bibliographystyle{elsarticle-num}
\bibliography{bib}
\end{document}